\documentclass[Runningheads]{llncs}
\usepackage[T1]{fontenc}
\usepackage{graphicx}
\usepackage[utf8]{inputenc}
\usepackage{algorithm}  
\usepackage{algpseudocode}  
\usepackage{amsmath}  

\usepackage[title]{appendix}
\usepackage{float}

\usepackage[marginal]{footmisc}

\usepackage{color}
\usepackage{subfigure}

\begin{document}
\title{Hierarchical Expert Prompt for Large-Language-Model:
An Approach Defeat Elite AI in TextStarCraft II for the First Time}
%
%
\author{
Zongyuan Li\inst{1,*} \and
Chang Lu\inst{1,*} \and
Xiaojie Xu\inst{1} \and
Runnan Qi\inst{2} \and
Yanan Ni\inst{2} \and
Lumin Jiang\inst{2} \and
Xiangbei Liu\inst{1} \and
Xuebo Zhang\inst{1} \and
Yongchun Fang\inst{1} \and
Kuihua Huang\inst{2,**} \and
Xian Guo\inst{1,**}
}
\authorrunning{Zongyuan Li et al.}
%
\institute{College of Artificial Intelligence, Nankai University, Tianjing, China \\
\email{\{2120230524, 2111533, 2120220490, 2120230527\}@mail.nankai.edu.cn}\\
\email{\{fangyc, zhangxuebo, guoxian\}@nankai.edu.cn}
\\
\and
Laboratory for Big Data and Decision, National University of Defense Technology, Changsha, China\\
\email{\{qirunnan13579, niyanan, jlm\_mz, khhuang\}@nudt.edu.cn}
}
\maketitle              
\footnote{* Zongyuan Li and Chang Lu contributed equally
to this work.\\ ** Corresponding author.}
\vspace{-0.5cm}
\begin{abstract}
Since the emergence of the Large Language Model (LLM), LLM has been widely used in fields such as writing, translating, and searching. However, there is still great potential for LLM-based methods in handling complex tasks such as decision-making in the StarCraft II environment. To address problems such as lack of relevant knowledge and poor control over subtasks of varying importance, we propose a Hierarchical Expert Prompt (HEP) for LLM. Our method improves the understanding of game situations through expert-level tactical knowledge, improving the processing quality of tasks of varying importance through a hierarchical framework. Our approach defeated the highest level (Elite) standard built-in agent in TextStarCraft II for the first time and consistently outperformed the baseline method\cite{Baseline} in other difficulties. Our experiments suggest that the proposed method is a practical solution for tackling complex decision-making challenges. The replay video can be viewed on \textcolor{blue}{https://www.bilibili.com/video/BV1uz42187EF} and \textcolor{blue}{https://youtu.be/dO3PshWLV5M}, and our codes have been open-sourced on \textcolor{blue}{https://github.com/luchang1113/HEP-LLM-play-StarCraftII}.
\end{abstract}
\section{Introduction}
StarCraft II, known for its complexity, is commonly utilized as a testing ground for decision-making algorithms, serving as a representative platform for complex decision-making tasks. In 2017, DeepMind, in collaboration with Blizzard, developed the StarCraft II game into StarCraft II Learning Environment\cite{SC2LE} (SC2LE). Two years later, AlphaStar\cite{AlphaStar}, an algorithm based on reinforcement learning (RL), defeated the world champion of StarCraft II at BlizzCon, proving that intelligent algorithms first surpass expert-level decision-making abilities in complex tasks. In 2022, an open-source algorithm DI-star\cite{DIstar} was proposed, reaching the highest segmented competition among human masters. Nevertheless, all these methods use a large amount of trajectory data and undergo long-term training on powerful computing devices. At the same time, there is often a lack of trajectory data to train algorithms in other tasks, which limits the deployment of RL-based algorithms in real-world decision-making problems. 

In November 2022 and April 2023, OpenAI released ChatGPT\cite{ChatGPT} and GPT-4\cite{GPT-4}, which enables computers to understand natural language and answer questions. Under this circumstance, the Institute of Automation, Chinese Academy of Sciences, constructed TextStarCraft II\cite{Baseline}, an environment that takes text information as input and output, providing a decision-making environment with enough complexity in LLM research. At the same time, they also proposed a decision framework called Chain of Summary (CoS) as a benchmark for decision LLM in TextStarCraft II. Compared to RL-based methods, this method can directly analyze observed information to make decisions without further training, showcasing potential in LLM-based decision-making. 

However, the LLM equipped with CoS defeated only the build-in agents ranging from VeryEasy to Harder. Yet it failed to secure a single victory in 12 matches against the bots set to the VeryHard difficulty. The following issues are associated with this method: (1)Due to the direct use of StarCraft II knowledge obtained from pre-train, LLM has a weak understanding of resource management, military development, and technology upgrades. (2)The situation analysis and suggestions are relatively rough and cannot effectively handle fine-grained information, ultimately leading to decisions that are vulnerable to mistakes in some critical problems. 

To resolve these challenges, we propose the Hierarchical Expert Prompt for LLM, which introduces expert tactic knowledge to LLM and improves the quality of processing key issues such as the construction of Nexus and Mineral-Gas management. Contrasted with Imitation Learning, injecting textual knowledge aligns with the customary approach to knowledge representation. It also circumvents the need for the imitation phase which demands a lot of trajectory data and computational resources. This approach effectively incorporates high-quality external knowledge into the model at a minimal expense.

In our experiments, we tested our approach in high-difficulty combat scenarios, analyzed time-varying data curves such as resource and supply, and conducted ablation studies on the Expert Tactic Module and the Hierarchical Decision Module. Results show that LLM with HEP can not only defeat VeryHard opponents with a high winning rate but also, for the first time, defeat Elite AI in this environment. Ablation studies show that the two proposed modules are indispensable in improving the quality of LLM-based decision-making.  
\section{Related Works}
\subsection{StarCraft-II Decision Environment}
Since the inception of StarCraft II, it has been receiving a great deal of
attention. With its characteristics of huge state space and action space, multi-agent, incomplete information, long time series, real-time, and good software support, it is recognized as one of the best environments for studying decision-making algorithms. 

In 2017, the SC2LE learning environment was proposed, followed by the proposal of the SMAC environment\cite{SMAC} two years later. However, these environments do not provide textual observations and cannot recognize textual instructions, limiting the ability to interact with LLMs.

\begin{figure}
\centerline{\includegraphics[width=320pt]{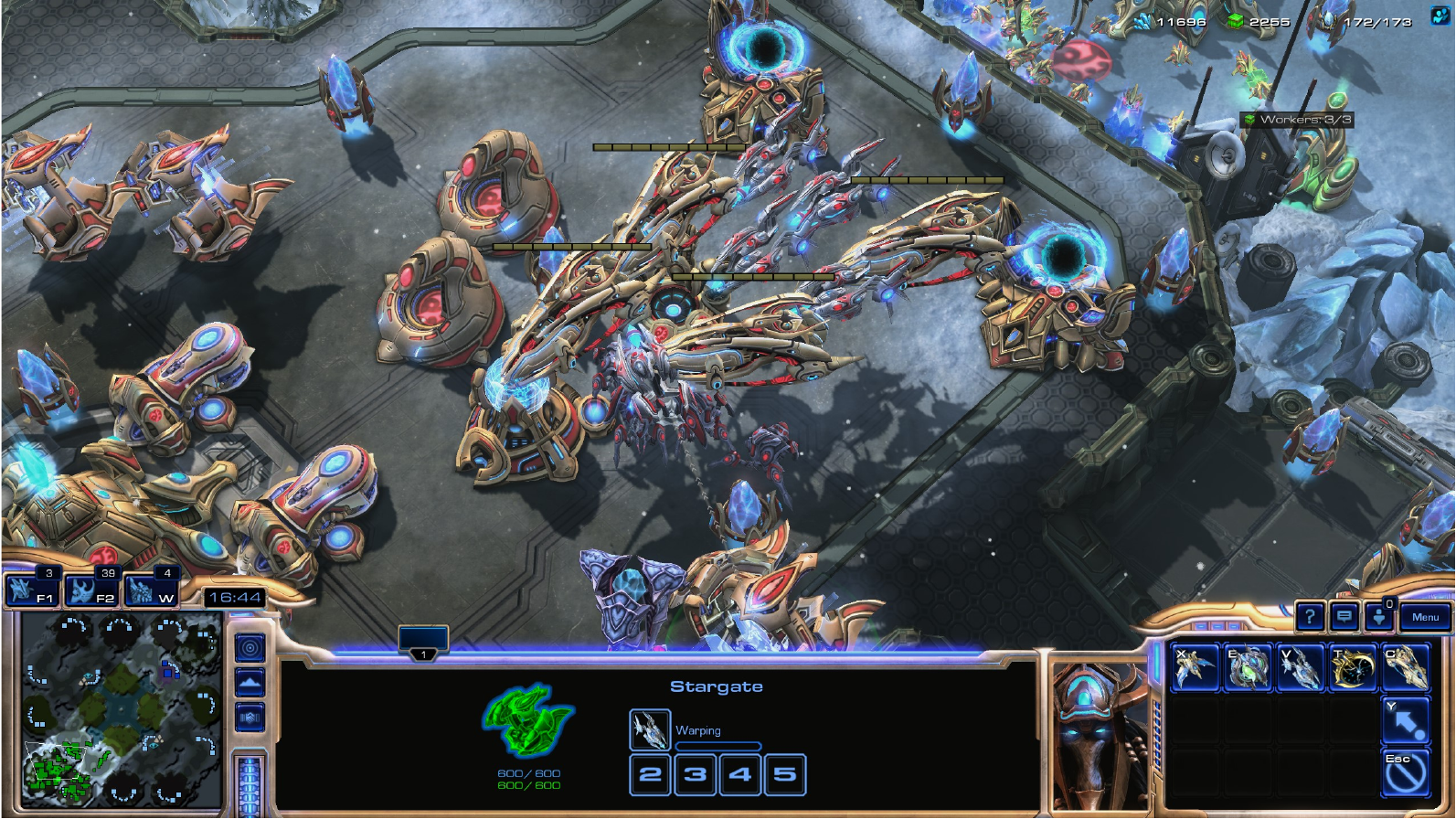}}
\caption{\textbf{StarCraft II.} In this decision-making environment, players need to control units, collect resources, build and upgrade technology, and confront opponents with incomplete observation information, making it one of the most complex decision-making environments.} 
\end{figure}

To provide an environment for LLM-based decision-making methods, TextStarCraft II was proposed. This environment separates macro decision-making and micro-operations, liberates LLM from high-speed micro-operations by handing over these operations automatically, and enables LLM to focus on macro decision-making. In interactive interfaces, the environment organizes the observed information into textual summaries and can recognize textual instructions, providing favorable conditions for LLM-based methods.

\subsection{Decision Making in Multi-Agent Games}

Multi-agent games include two directions of research. One is the study of a simple game with many agents, such as 27Marine-vs-30Marine task, controlling a group of agents to achieve maximum combat capability in the SC2LE . The second is the study of complex decision tasks with fewer agents, such as playing Go\cite{AlphaGo}, Texas Poker\cite{Alphaholdem}, Mujoco Football Game\cite{MujocoFootball}, and complete battle in StarCraft II. 

In the study of the first direction, many multi-agent reinforcement learning algorithms have emerged, such as value decomposition (VD) methods such as VDN\cite{VDN}, Qmix\cite{QMIX}, Weighted QMIX\cite{WQMIX} and so on, and centralized training decentralized execution (CTDE) methods such as MAPPO\cite{MAPPO} and MADDPG\cite{MADDPG}. These algorithms focus on solving the credit assignment problem of agents in a team, avoiding lazy agent problem, and aiming at maximizing the collaborative efficiency of the group of agents.

In the study of the second direction, algorithms such as AlphaGo\cite{AlphaGo}, AlphaStar, and AlphaHoldem\cite{Alphaholdem} showcase the ability to surpass top human players in complex multi-agent games. As a milestone in StarCraft II decision-making, AlphaStar uses two-stage training to acquire decision-making abilities beyond humans, first training on expert trajectories for about two weeks, then applying to more than one month of reinforcement learning to obtain higher abilities. 

However, all these algorithms mentioned above face the same problem: the demand for computing resources and training time. Some also require a large amount of high-quality trajectory data for imitation. These problems hinder further development of these methods and may also lead to the Sim-to-Real problem when migrating to other tasks.

\subsection{Large Language Model}
LLM usually refers to a language processing network with billions of parameters. In November 2022, OpenAI released the LLM-based chat application ChatGPT. In April 2023, OpenAI released GPT-4, which further improved the performance of LLM with more parameters and training data. After that, more and more LLM emerge, such as Claude-2\cite{Claude-2} and 3\cite{Claude-3}, LLAMA-2\cite{LLAMA-2} and 3\cite{LLAMA-3}.

LLMs have demonstrated remarkable skills in handling textual data, yet LLM-based decision-making systems have not been well advanced. In March 2024, an LLM enhanced with CoS achieved a milestone by securing a 50\% winning rate against Harder opponent in TextStarCraft II for the first time. However, this approach cannot defeat the VeryHard and Elite opponents. On their basis, we constructed Hierarchical Expert Prompt, introducing expert knowledge and enhancing the ability to deal with tasks with different priorities for LLM. For the first time, it defeated the TextStarCraft II agent under Elite difficulty, suggesting that the LLM-based method is a feasible solution to apply to decision-making tasks with greater complexity.
\section{Methods}

We propose Hierarchical Expert Prompt for StarCraft II decision-making. This approach comprises two primary components: the Expert Tactic Prompt (ETP) and the Hierarchical Decision Prompt (HDP). All of our prompts are fully displayed in Appendix A.

\vspace{-0.3cm}
\begin{figure}
\centerline{\includegraphics[width=320pt]{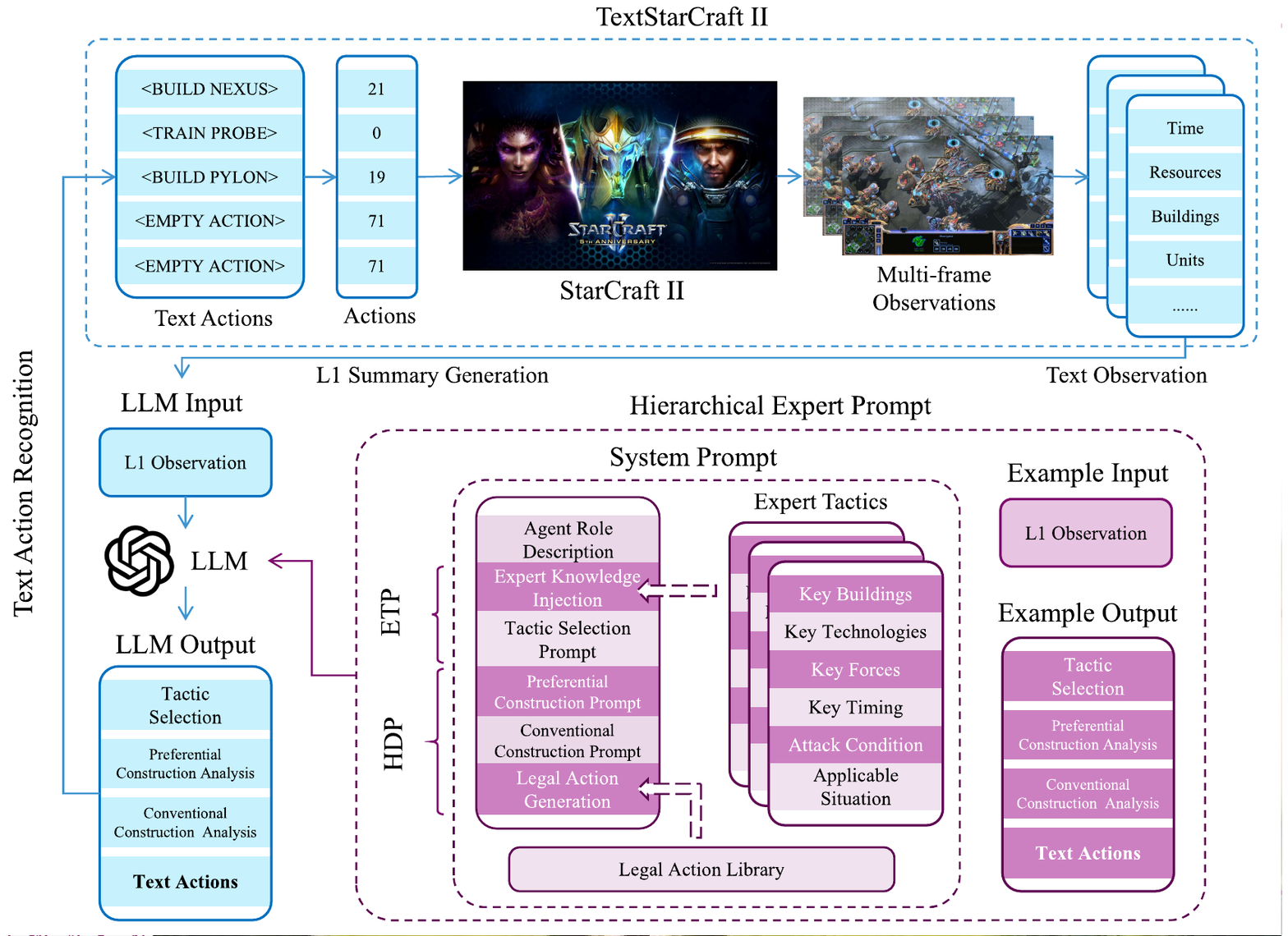}}
\caption{\textbf{Interacting with LLM: Hierarchical Expert Prompt Method in TextStarCraft II.} LLM takes the L1 summary\cite{Baseline} (a highly condensed text observation) as input, obtains knowledge from Hierarchical Expert Prompt, and generates text actions according to analyses.} \label{fig2}
\end{figure}

\vspace{-0.3cm}
As shown in Fig~\ref{fig2}, HEP consists of a system prompt, an example input prompt, and an example output prompt. In the part of the system prompt, we link the Role Prompt, ETP, HDP, and Legal Action Library together. LLM then absorbs knowledge from ETP and follows the hierarchical decision logic of HDP. The methodology for constructing the HEP is detailed in Algorithm 1, while the interaction process between LLM and environment is illustrated in Algorithm 2.


\subsection{Expert Tactic Prompt}

To implement expert-level tactic knowledge, we propose ETP, construct several classic tactics into an expert knowledge database, and inject them into the LLM as part of the system prompt. The LLM then assesses the situation, selects an appropriate tactic, makes decisions based on the chosen tactic, and gathers information. After all aspects are well developed, the LLM agent will launch a decisive attack to defeat the opponent.

\begin{algorithm}
\caption{Hierarchical Expert Prompt Generation}
\begin{algorithmic}

\Require Text observation $o_t$; Agent role prompt $p_r$; Expert Tactic Prompt $p_{ETP}$; Hierarchical Decision Prompt $p_{HDP}$; Legal Action Library $L$; Example input $p_i$ and output $p_o$. All the quantities above are in text form.
\Ensure LLM query message $m_t$

\Function{$Get\_HEP\_message$}{$o_t$}
    \State Initialize openai standard message object $m_t$
    \State Redefine operation '+' for text concatenating
    \State $m_t.append(p_r + p_{ETP} + p_{HDP} + L, role="system")$
    \State $m_t.append(p_i, role="user")$
    \State $m_t.append(p_o, role="assistant")$
    \State $m_t.append(o_t, role="user")$
    \State Return $m_t$
\EndFunction

\end{algorithmic}
\end{algorithm}

\subsubsection{Expert Tactic Knowledge Base}

We propose a standardized description for StarCraft II tactics. Each tactical description prompt consists of the following components: name, key buildings, key technologies, key forces, key timing, and applicable situation. Describing these parts clearly could make external knowledge more recognizable for LLM. The standardized tactic description can be found in Appendix A.

To test the ability to change tactics, we experimentally propose two classic Protoss tactics for LLM and construct the knowledge base: One is an early-game tactic that focuses on units like Zealots, creating a significant threat to the Zegling system of the Zerg race. The second focuses on carriers, aiming at delivering devastating strikes from the sky and achieving victory in the middle or later stage.

\subsubsection{Knowledge Injection and Tactic Switch}

To achieve knowledge injection, we incorporate the knowledge base into the system prompt, which is the most intuitive way to introduce expert knowledge to LLM. To enable tactical switching and enhance the ability of LLM to adapt to different situations, we introduce a tactical selection module before conducting situation analyses and delivering final actions. This module requires the LLM to select tactics based on the current situation and specify them in the <current tactic> field.

\subsection{Hierarchical Decision Prompt}


Hierarchical structure is often employed in decision-making problems. Inspired by hierarchical decision-making algorithms such as Decision Tree and Hierarchical Reinforcement Learning, we propose the Hierarchical Decision Prompt. We separate legal actions into two groups with different priorities and ask the LLM to choose actions following the hierarchical decision logic of HDP. The LLM first conducts the necessary analysis, determines whether there are any priorities, selects a set of actions that follow the hierarchical decision logic, and finally generates textual legal actions at the end of the output prompt. 



\begin{algorithm}
\caption{HEP-Agent Interaction in TextStarCraft II}
\begin{algorithmic}
\Require Environment Env (TextStarCraft II); Large Language Model LLM; Interaction frequency n; Default observation: $o_0$; Default action $a_0$.

\State Initilalize $o_t = o_0$ 
\State Initilalize $a_t = a_0$
\State Initilalize environment $Env.reset()$
\While{not $Env.is\_terminated()$}
    \If{$t \% n = 0$}
        \State $o_t = Env.get\_obs()$
        \State $m_t = Get\_HEP\_message(o_t)$
        \State $a_t = LLM.query(m_t)$
        \State $Env.step(a_t)$
    \Else
        \State $Env.step(a_0)$
    \EndIf
\EndWhile

\end{algorithmic}
\end{algorithm}

\subsubsection{Priority and Routine Task Analysis(Lower Layer)}

In StarCraft II, the construction of Nexus and Assimilator will greatly affect long-term development. When routine tasks such as training units and constructing other buildings will occupy funding for building Nexus and Assimilator. So, we separate actions into priority group and routine task group and ask the LLM to analyze relevant information in preferential constructing items (build Nexus and Assimilator) and conventional constructing items (economy development, technology development, military development, chronoboost, scouting, and whether developed well to launch the final attack). Specify <priority> field at the end of the priority analysis if Nexus or Assimilator should be built immediately.

\subsubsection{Selection of Action Group(Upper Layer)} 
 
To improve LLM's capabilities in dealing with tasks with different priorities, we set a hierarchical logic control text between situation analysis and action generation, instructing LLM to select an action group following the hierarchical logic. The hierarchical logic control text first asks the LLM to determine whether <priority> is None. If <priority> is None, the hierarchical logic text guides the LLM to generate actions according to routine task analysis. otherwise, guide the LLM to generate actions according to priority analysis and prevent generating actions of the other group to avoid unnecessary resource consumption (except training Probe and building Pylon).

\subsubsection{Legal Action Generating Prompt}

As the last part of HDP, the Legal Action Generating Prompt is indispensable in generating valid actions. We set the legal action library as part of the prompt, ask the LLM to choose actions from the library according to situation analysis. Finally, the generated legal actions will be recognized by the TextStarCraft II environment, converting to corresponding micro-operations and interacting with the StarCraft II software. 


\clearpage
\section{Experiments}

All the experiments run on the TextStarCraft-II environment. To compare with the baseline results\cite{Baseline}, we used exactly the same setting as in Ref.1.

The content of this chapter is arranged as follows. In section 4.1, we tested HEP-Agent fighting against opponents from Hard (Level-4) to Elite (Level-7), calculated the winning rates, and compared the costs of two methods. In section 4.2, we visualized the changes in population and resources, made a detailed analysis of the battle process. In the final section, we conducted ablation studies on proposed modules. Experiments results show that (1)Our method significantly improves the decision-making ability of LLM at a reasonable cost. (2)Both the proposed modules are essential to our method.

\subsection{Evaluation of the Hierarchical Expert Prompt method}

In this section, we compared the winning rates and costs of our method with the baseline method. In our experiments, agents with GPT-3.5 back-end played Protoss in TextStarCraft-II against Zerg of varying difficulties. As shown in Table~\ref{tab1}, our method increased the winning rate by 16\% and 25\% on Harder and VeryHard difficulties and managed to defeat VeryHard opponents with a 75\% winning rate. It is worth noting that, our method even obtained victories from Elite opponents, indicating that LLM can defeat the highest level standard build-in AI without further training.

\begin{table}
\caption{\textbf{Winning Rates in Difficulties Ranging from Level-4 to Level-7.}}\label{tab1}
\begin{center}
\vspace{-0.3cm}
\begin{tabular}{p{2cm} p{2.5cm} p{2.5cm} p{2.5cm} p{2.0cm} }
\hline
Method & \multicolumn{4}{c}{Difficulty Level} \\
\hline
 & Hard & Harder & VeryHard & Elite \\
\hline
Baseline & 21/25 (84\%) & 7/14 (50\%) & 0/12 (0\%) & TBD \\
Ours & 12/12 (100\%) & 9/12 (75\%) & 9/12 (75\%) & 3/12 (25\%) \\
\hline
\end{tabular}
\end{center}
\end{table}

\vspace{-1.0cm}
\begin{table}
\caption{\textbf{API-Calling Cost of the Two Methods in Time and Tokens.}}\label{tab-cost}
\begin{center}
\vspace{-0.3cm}
\begin{tabular}{p{2.0cm} p{2.0cm} p{2.5cm} p{2.5cm} p{2.5cm}}
\hline
Method & Time & Total Token & Prompt Token & Output Token\\
\hline
Baseline & 12.43s & 3209 & 2559 & 650\\
Ours & 13.78s & 4372 & 3723 & 649 \\
\hline
\end{tabular}
\end{center}
\end{table}
\vspace{-0.5cm}

We also tested the API-calling consumption of both methods in time and tokens. As shown in Table~\ref{tab-cost}, although our method increased consumption by 10.86\% in time and 36.24\% in total tokens, two level breakthroughs were obtained and winning rates increased in all tested difficulties from Hard to Elite, which means that our method significantly improves the decision-making ability of LLM at a reasonable cost. 

\begin{figure}
    \centering
    \subfigure[The LLM agent started to construct the second Nexus at about 1:16. ]{\includegraphics[width=0.49\textwidth]{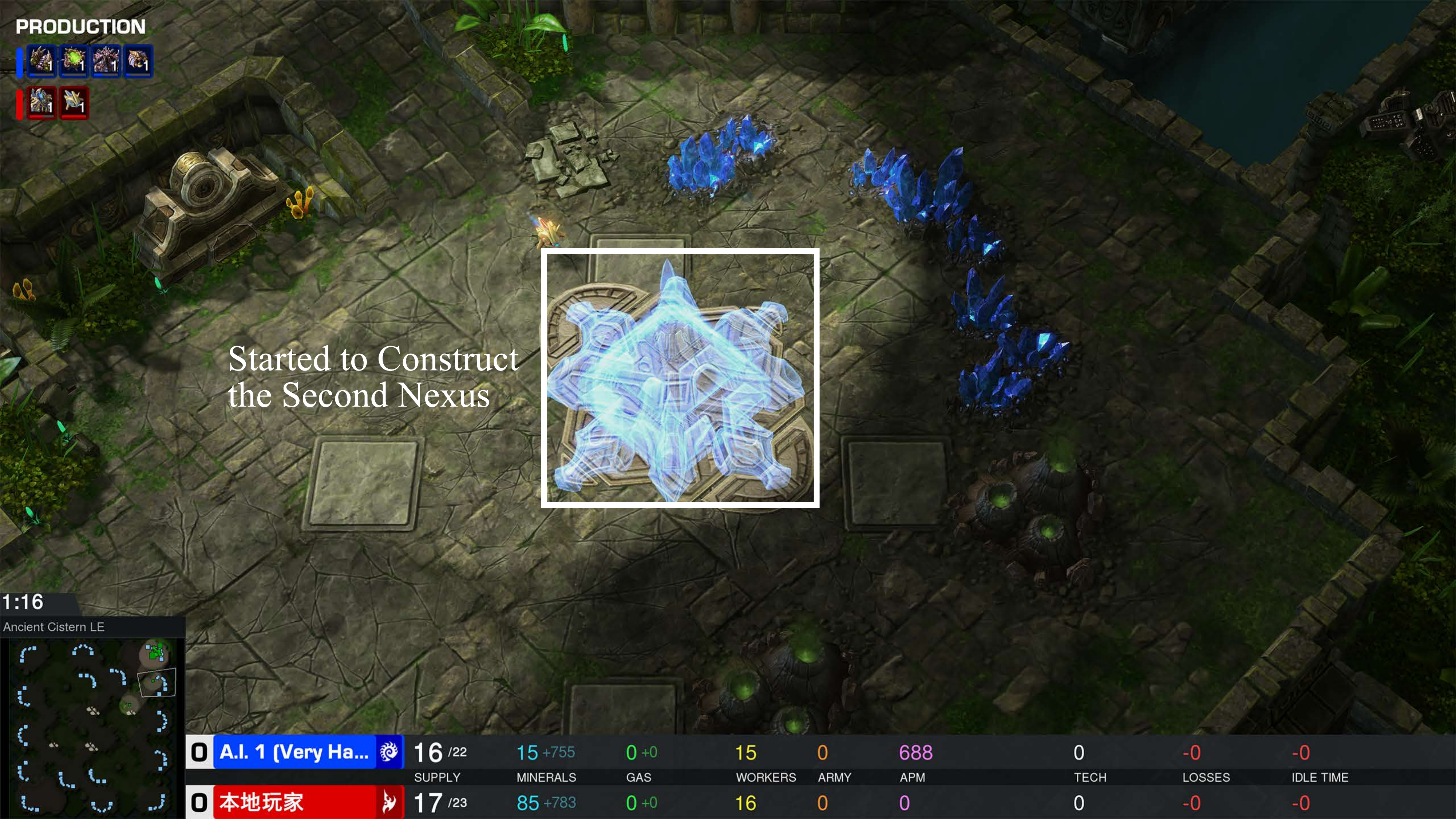}}
    \subfigure[Our Probe performed scouting, arrives at enemy's base before 2:00. ]{\includegraphics[width=0.49\textwidth]{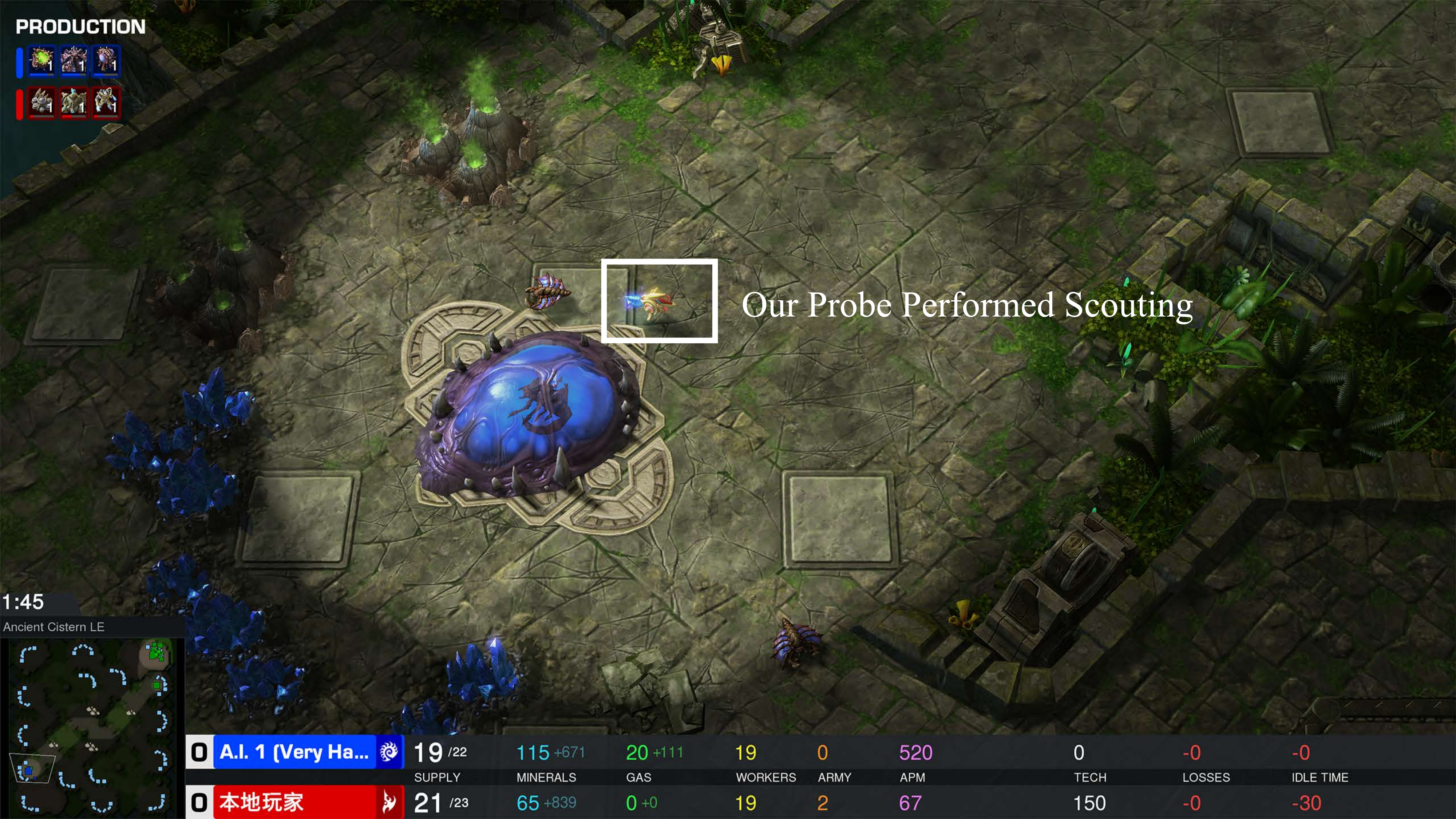}} \\
    \subfigure[Two carriers were training in stargates synchronously at 8:56. ]{\includegraphics[width=0.49\textwidth]{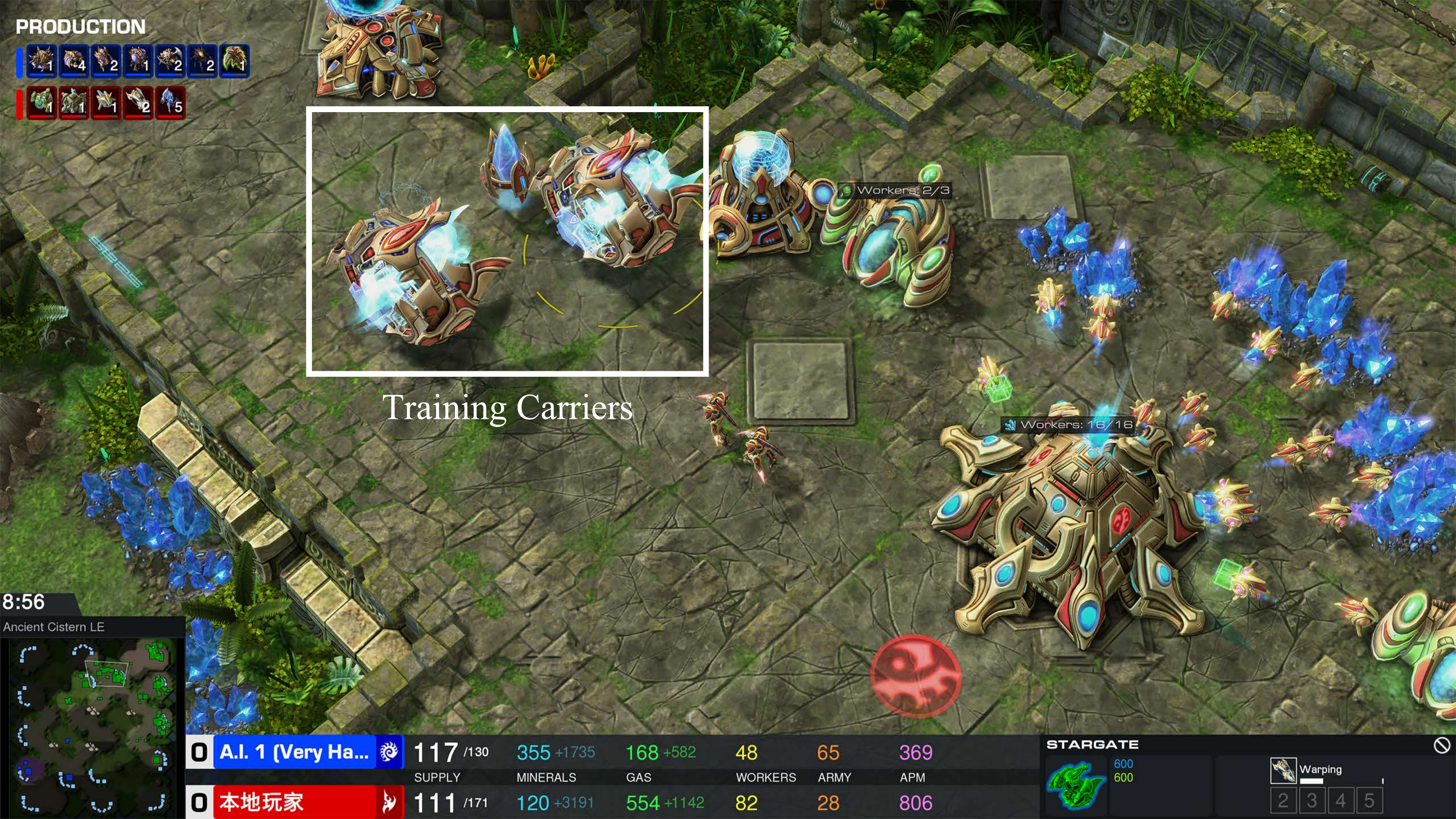}}
    \subfigure[Ground force suffered damage and created opportunities for killing anti-air units.]{\includegraphics[width=0.49\textwidth]{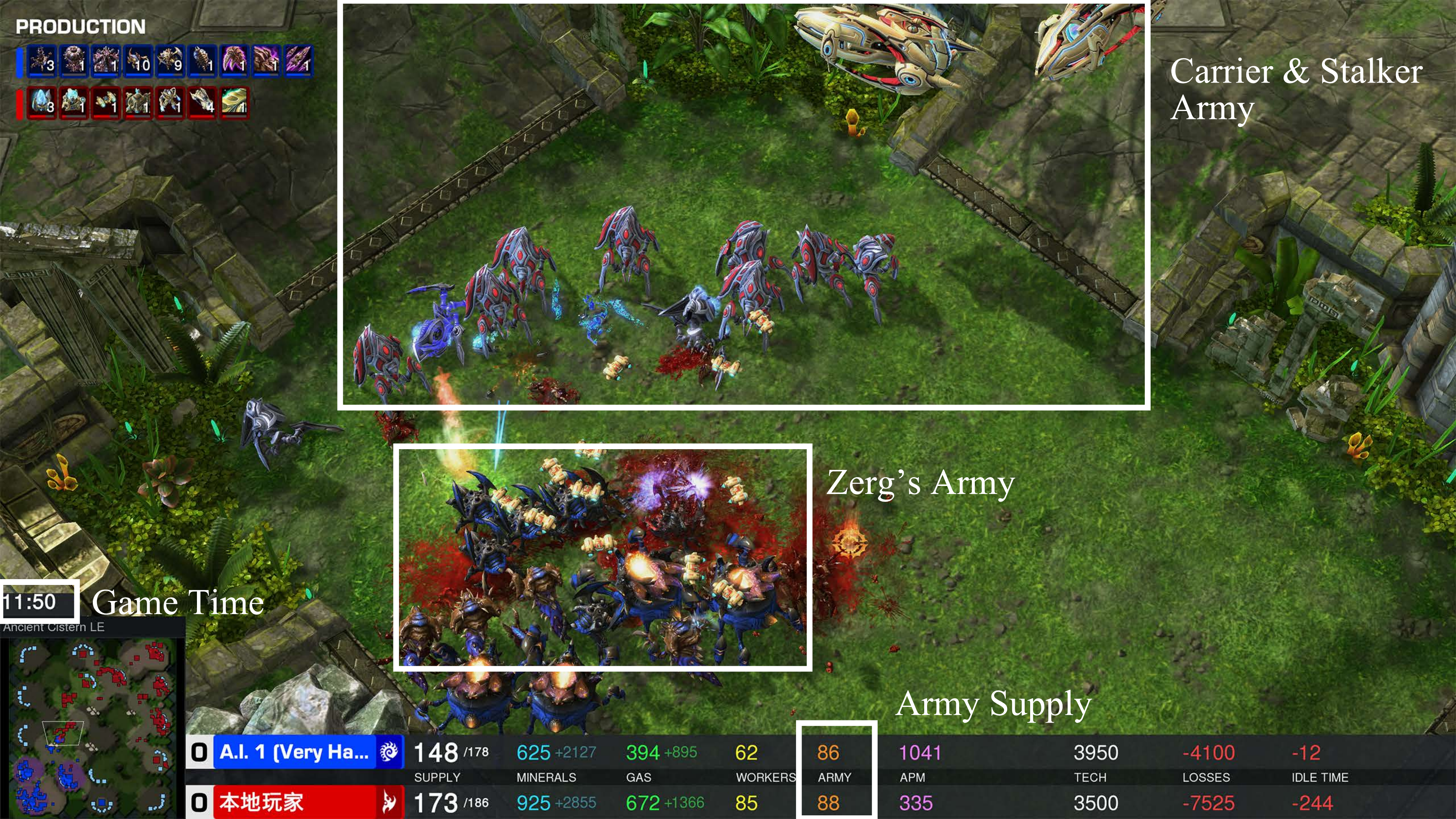}} \\
    \subfigure[All required technologies of carrier tactic have been completed in the game.]{\includegraphics[width=0.49\textwidth]{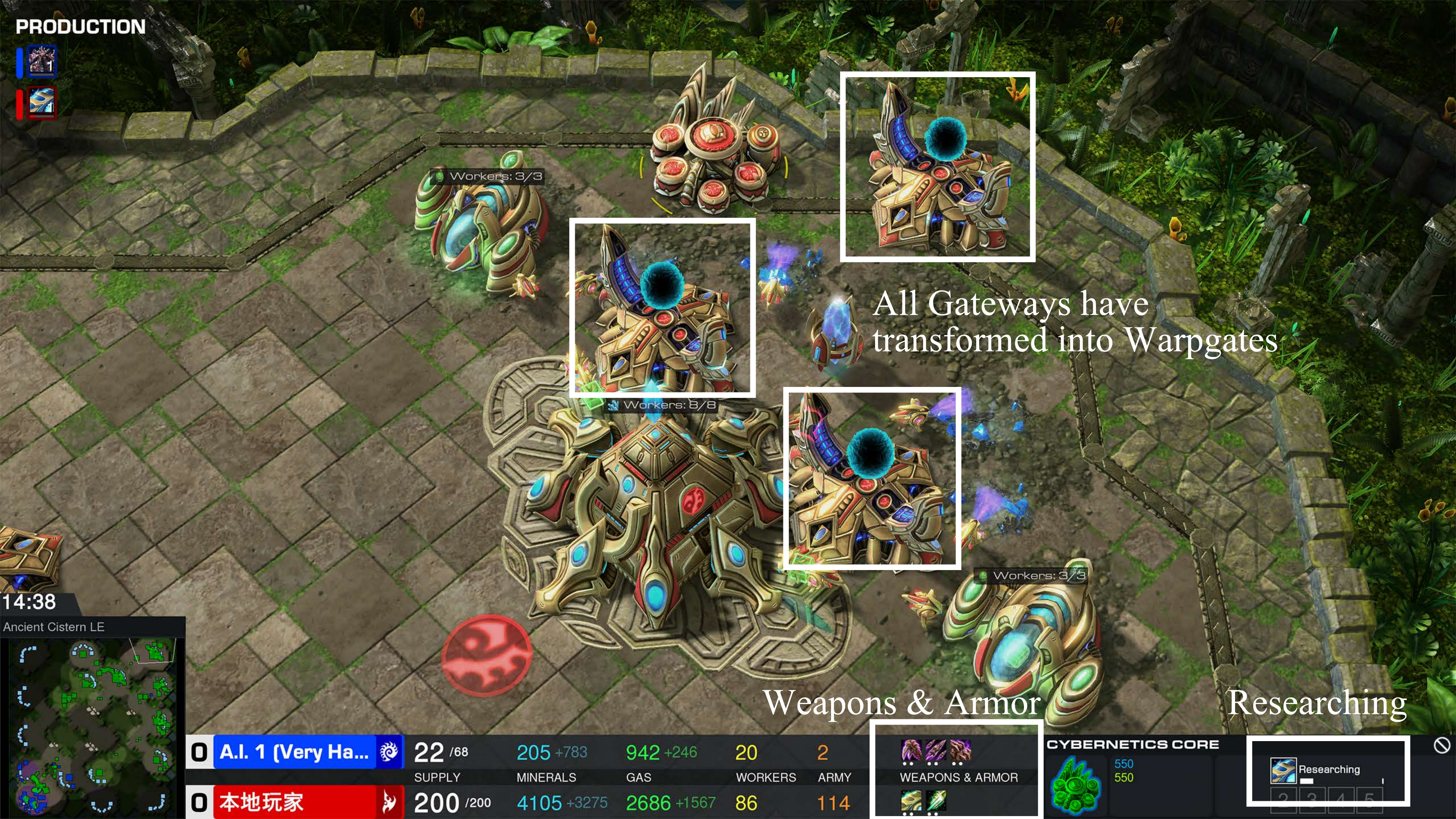}}
    \subfigure[Carriers and stalkers occupied the highland and destroyed the zerg's base.]{\includegraphics[width=0.49\textwidth]{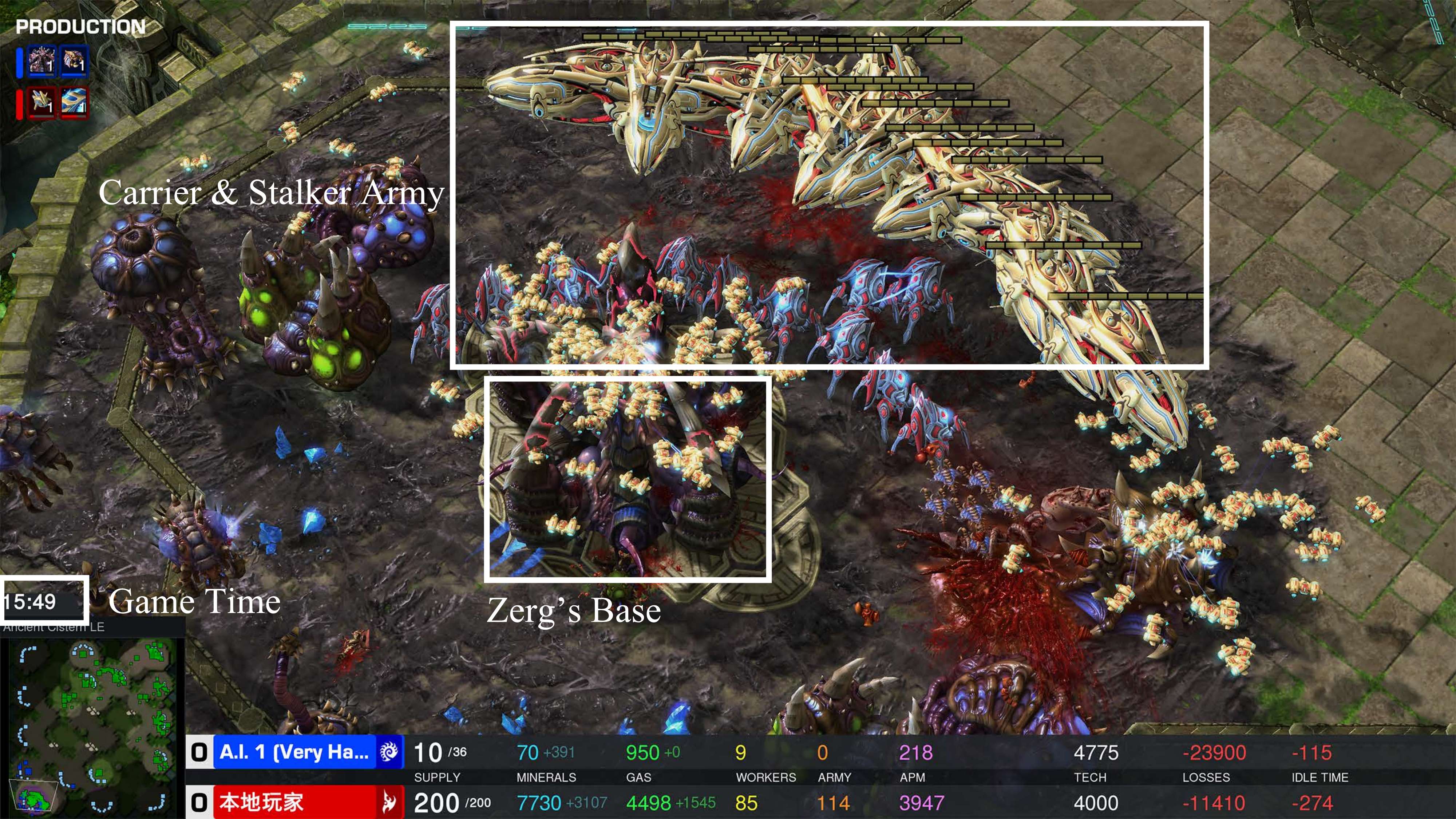}} \\
    \caption{\textbf{Screenshots of Game Replay Against VeryHard Opponent.}}
    \label{fig:screenshots}
\end{figure}

\clearpage
We also displayed some screenshots of the replay. As shown in Fig~\ref{fig:screenshots}, our HEP-Agent starts building the second nexus within 2 minutes, scouting as soon as possible, training powerful units at the middle stage of the game, completing required technologies, defeat the opponent's military forces and finally destroy the base. More details can be viewed on \textcolor{blue}{https://www.bilibili.com/video/BV1uz42187EF} and \textcolor{blue}{https://youtu.be/dO3PshWLV5M}.

\clearpage
\subsection{Detailed Analysis}

\vspace{-0.5cm}
To figure out how our method defeated VeryHard AI, we visualized key data such as mineral and gas storage, estimation of total collected resources, worker and army supply, the composition of troops, and technological upgrades during the game. All the data visualized here were collected in games of difficulty level-6, and results of experiments in other difficulties can be found in Appendix B. 

\vspace{-1.2cm}
\subsubsection{Economy}

\begin{figure}[H]
\centerline{\includegraphics[width=\textwidth]{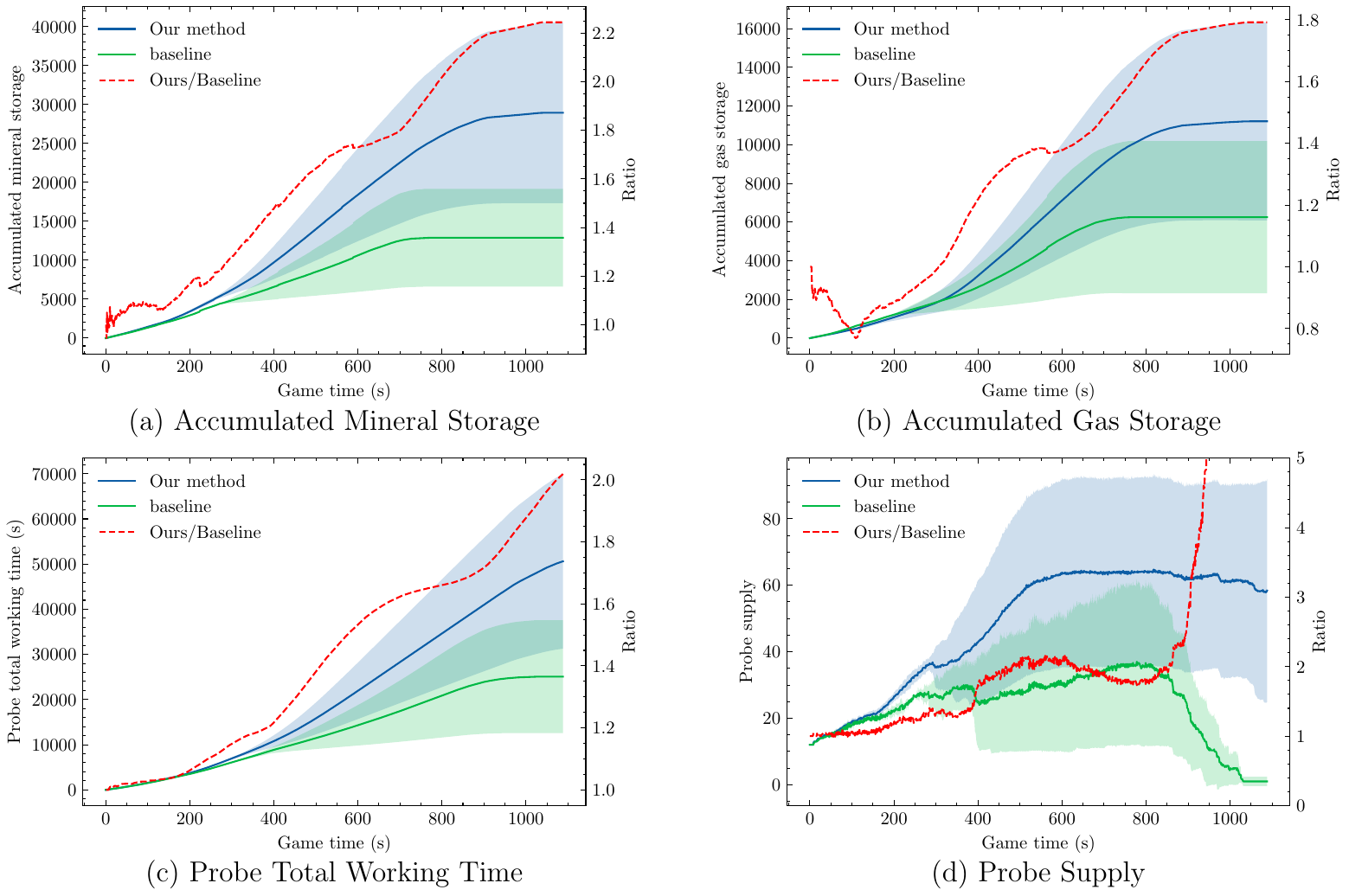}}
\caption{\textbf{Resource Data.} (a)(b) In the first 8 minutes, our method focused on collecting minerals and the amount of gas was low, after 8 minutes, our method started to collect a lot of gas to provide resources for building the army in the later stages. (c)(d) Our method is able to train more Probes compared to the baseline. } \label{fig3}
\end{figure}

\begin{figure}[H]
\centerline{\includegraphics[width=\textwidth]{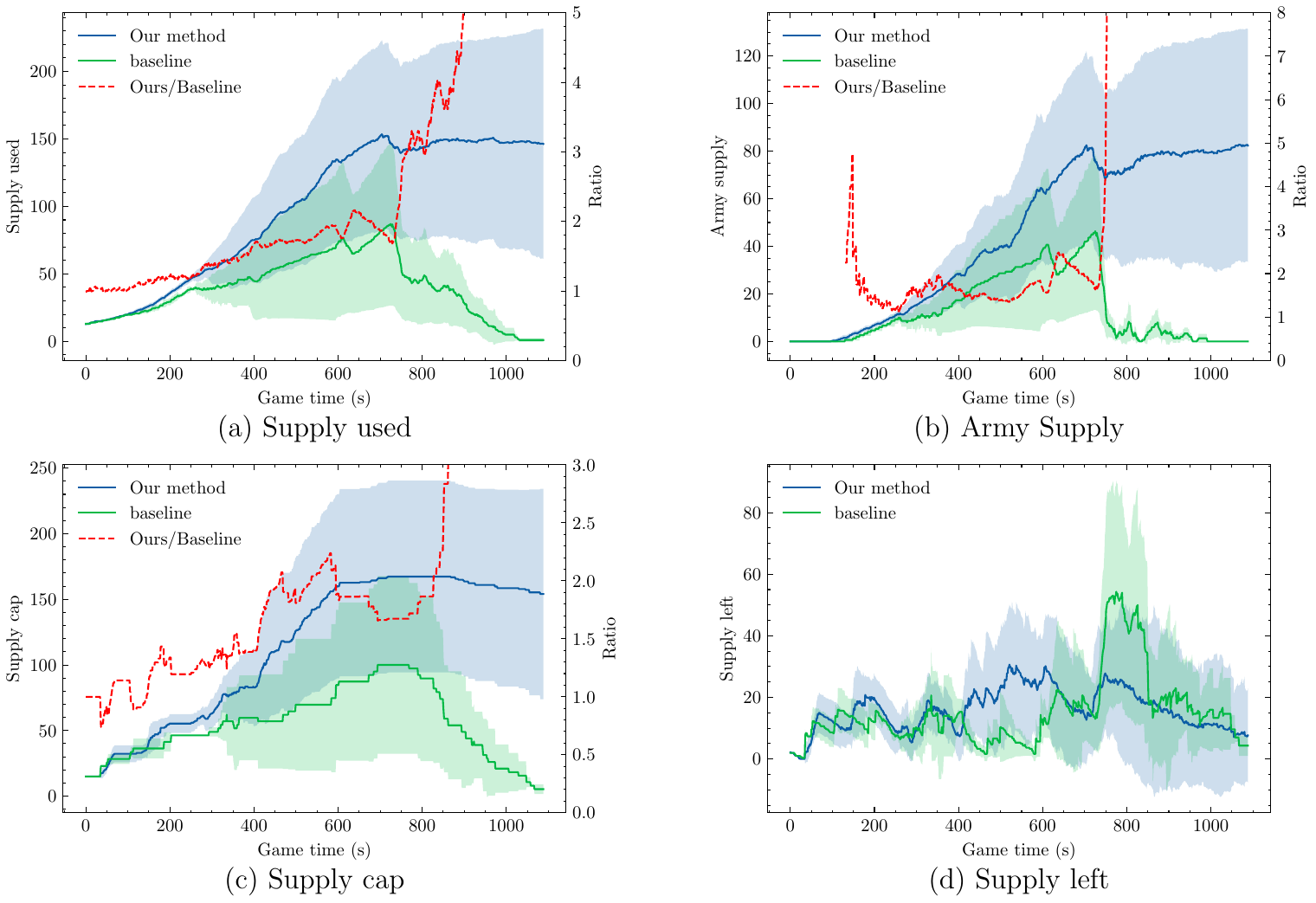}}
\caption{\textbf{Supply Data.} (a)(b) Our method is able to use more supply compared to baseline, importantly, the army supply has a larger increase in the early game, makes it possible to defend the opponent's early attack and scouting. (c)(d) Our method continuously built Pylons to prevent running out of supply. During 400s to 600s, the baseline method struggled in increasing supply, while our method kept the supply increasing continuous.} \label{fig4}
\end{figure}

We visualized some economic-related data of both methods during the game, including resources and supplies. As shown in Figure~\ref{fig3} and Figure~\ref{fig4}, our method managed to collect 20\% more mineral than the baseline in 5 minutes and 40\% more gas in 8 minutes
. Meanwhile, our method can build more Pylons and train more units at a corresponding speed. 
All of the results above indicate that our method provides a significant economic improvement over the baseline method, and the improvement increases over time. 

\subsubsection{Technology} 

Technology plays a decisive role in the result of StarCraft battles. We selected a typical battle for both the baseline method and ours and recorded the status of technology research in Table~\ref{tab-tech}. Results show that our method completed nearly twice the technological upgrade, enabling our units to have stronger combat capabilities.

\begin{table}
\caption{Technology update status in different game time}\label{tab-tech}
\begin{center}
\vspace{-0.5cm}
\begin{tabular}{p{2.5cm} p{5cm} p{5cm}}
\hline
Time & Baseline & Ours\\
\hline
8min & Warpgate, 100\% & Warpgate, 100\% \\
     &  & Protoss-air-weapon-level-1, 83\% \\
\hline
16min & Warpgate, 100\% & Warpgate, 100\% \\
      & Protoss-air-weapon-level-1, 100\% & Protoss-air-weapon-level-2, 100\% \\
      & Protoss-air-armor-level-1, 100\% & Protoss-air-armor-level-2, 93\% \\
\hline
\end{tabular}
\end{center}
\end{table}

\subsubsection{Military}  

\begin{figure}[H]
\centerline{\includegraphics[width=\textwidth]{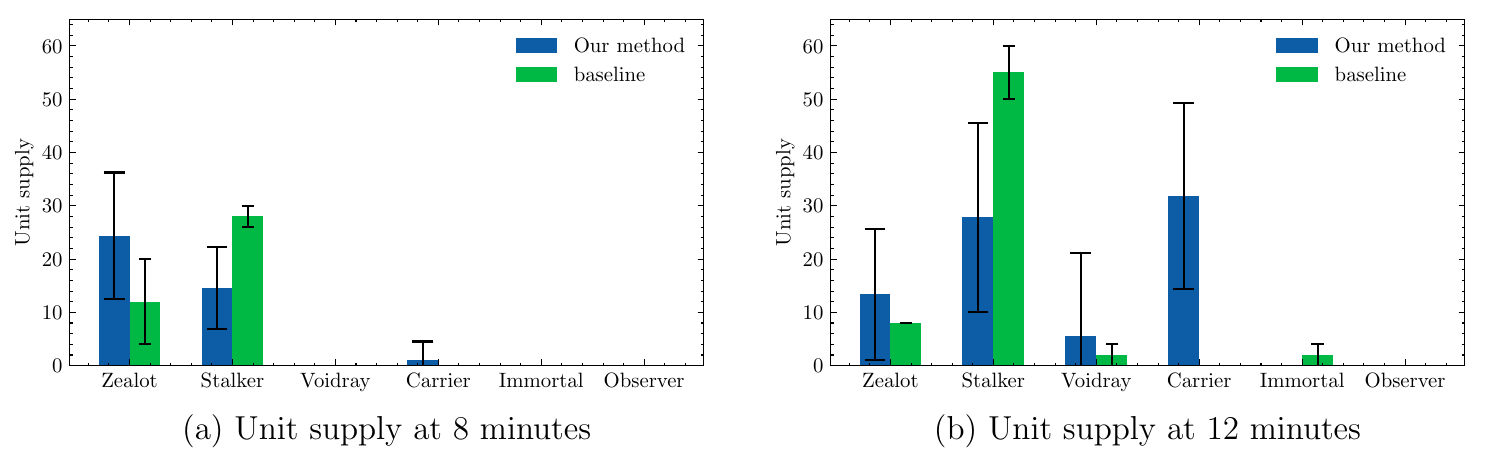}}
\caption{\textbf{Army Composition.} Unit supplies count at different points in time. Notably, baseline only trained Zealots and Stalkers as major combatant, our method is able to train Carriers to follow the tactics and achieve a stronger military force to defeat enemy. } \label{fig_unit_count}
\end{figure}

To compare the military strength of our method with the baseline method's, we visualized unit supplies at different time points. As shown in Figure~\ref{fig_unit_count}, for the first 8 minutes, both our method and baseline trained Zealot and Stalker to defend against enemy attack and scouting in the early stage. After 8 minutes, our method trained Carriers as the major military force, while baseline still trained weak Stalkers
The above results show that our method can build a stronger army 
to defeat the higher-level built-in AI.
\vspace{-0.5cm}

\vspace{-0.5cm}

\vspace{-0.5cm}
\subsubsection{Tactic}

\begin{figure}[H]
\centerline{\includegraphics[width=\textwidth]{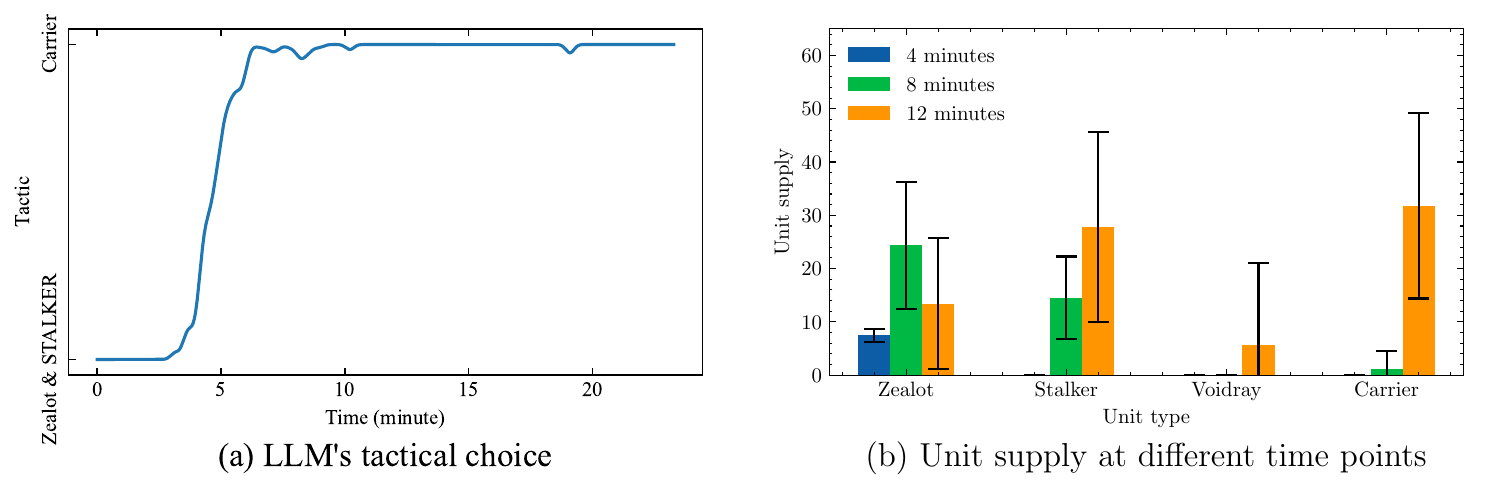}}
\caption{\textbf{Tactic Selection.} (a) Before 4 minutes, LLM chose Zealot \& Stalker tactic, after that LLM made a tactical switch to Carrier tactic, which followed the 'applicable conditions' in tactical prompt. (b) In the early stage of the game, LLM chooses Zealot \& Stalker tactic and trains zealots and stalkers, then, during 4 to 6 minites, LLM performs tactical switch and changes <Current Tactic> to Carrier tactic, constructs fleet beacon by the tactical requirements and starts training Carriers at 8 minutes.} 
\label{fig5}
\end{figure}

To verify that the LLM followed the instructions of the ETP module, we collected all LLM's outputs and extracted the tactics chosen by the LLM, and we also collected unit supply at 4, 8, 12, and 16 minutes to analyze the execution of the tactical switch. The results in Figure~\ref{fig5} show that the LLM followed the instructions of the expert tactic knowledge well, chose the correct tactic, built the army according to it, and ultimately wined the game. 
\vspace{-0.5cm}

\subsection{Ablation Study}

To prove the effectiveness of our proposed ETP and HDP, we set up two sets of ablation experiments and visualized some key data. Each group of ablation experiments performed three games against the VeryHard opponent. The prompts used in the ablation experiments can be found in Appendix A.2 and more detailed data can be found in Appendix B. 

The results in Figure 7 show that only a complete HEP can develop the economy, build a strong army with many carriers, and eventually win the game, suggesting that both the ETP and the HDP module are necessary parts of our method. Without EDP, LLM didn't know it needed to train carriers in the later stage of the game and could not build up the military strength thus losing the game. Without HDP, LLM has been slower to develop its economy, especially in the gas collection, making it impossible to train carriers and lose the game. 

\begin{figure}
\centerline{\includegraphics[width=\textwidth]{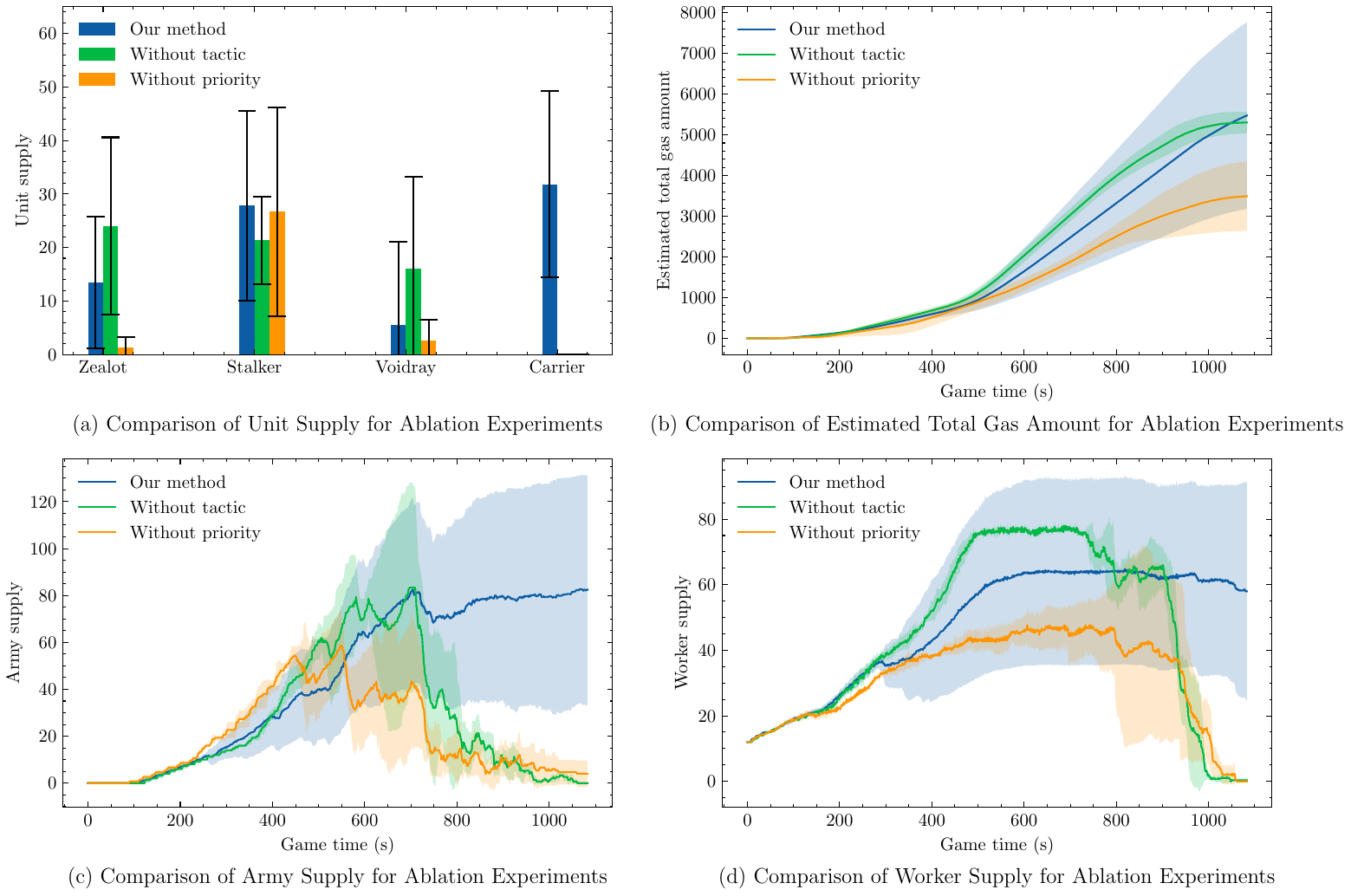}}
\caption{\textbf{Ablation Study Results.}  Results above shows the impact of two proposed modules of our method. (a)(b) Without tactics, LLM did not know how to train Carriers even collected a lot of gas, while without priority module, there is not enough gas to train Carriers in the later stage of the game. (c)(d) Both the two ablation groups can not survival form the battle at about 800s, lost almost all of its units after 1000s. } \label{fig_ablation_study}
\end{figure}


\section{Conclusion}
In this paper, we introduce HEP for decision-making LLM, inject textual knowledge into LLM, and improve the agent's ability to deal with tasks of varying importance. In experiments, results show that our method significantly improves the winning rates, and both proposed modules are indispensable. Most importantly, we prove that LLM can defeat Elite AI without any further training, revealing the great potential of LLM in dealing with complex decision-making problems.

%


\appendix


\clearpage
\section*{Apendix A. All Prompt}
\setcounter{figure}{0}
\renewcommand{\thefigure}{A\arabic{figure}}

\subsection*{A.1 Hierarchical Expert Prompt with Example input and output}
\textbf{A.1.1 System Prompt}

\begin{figure}
\centerline{\includegraphics[width=\textwidth]{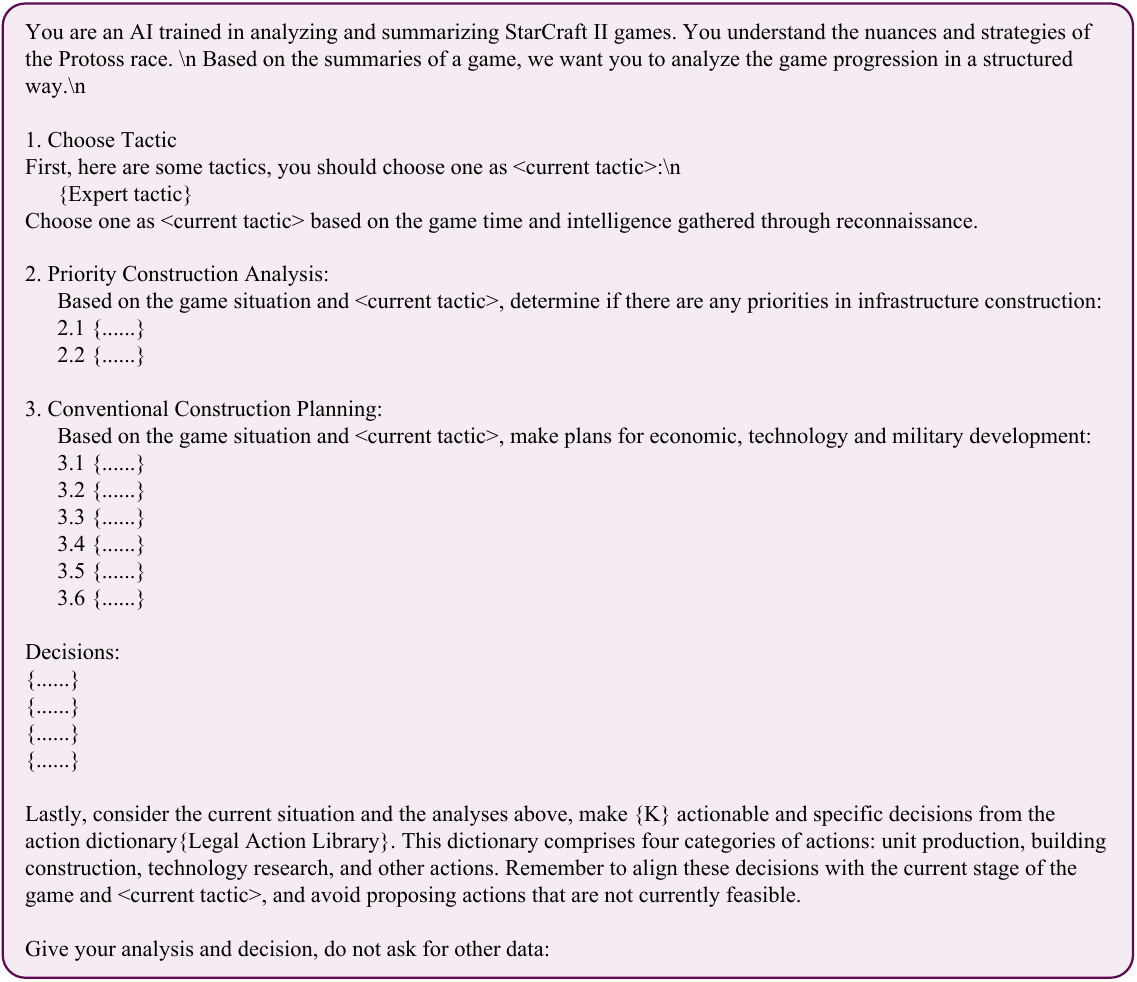}}
\caption{\textbf{General Structure of System Prompt.} Our system prompt consist of five parts: Agent role description, Tactic Selection (Choose Tactic), Preferential Construction Analysis (Priority Construction Analysis), Conventional Construction  Analysis (Conventional Construction Planning), Legal action generation (Decisions and the Legal Action Library below). Details of these parts can be viewed in the following pages, from Fig.A2 to Fig.A5.} 
\end{figure}

\begin{figure}
\centerline{\includegraphics[width=\textwidth]{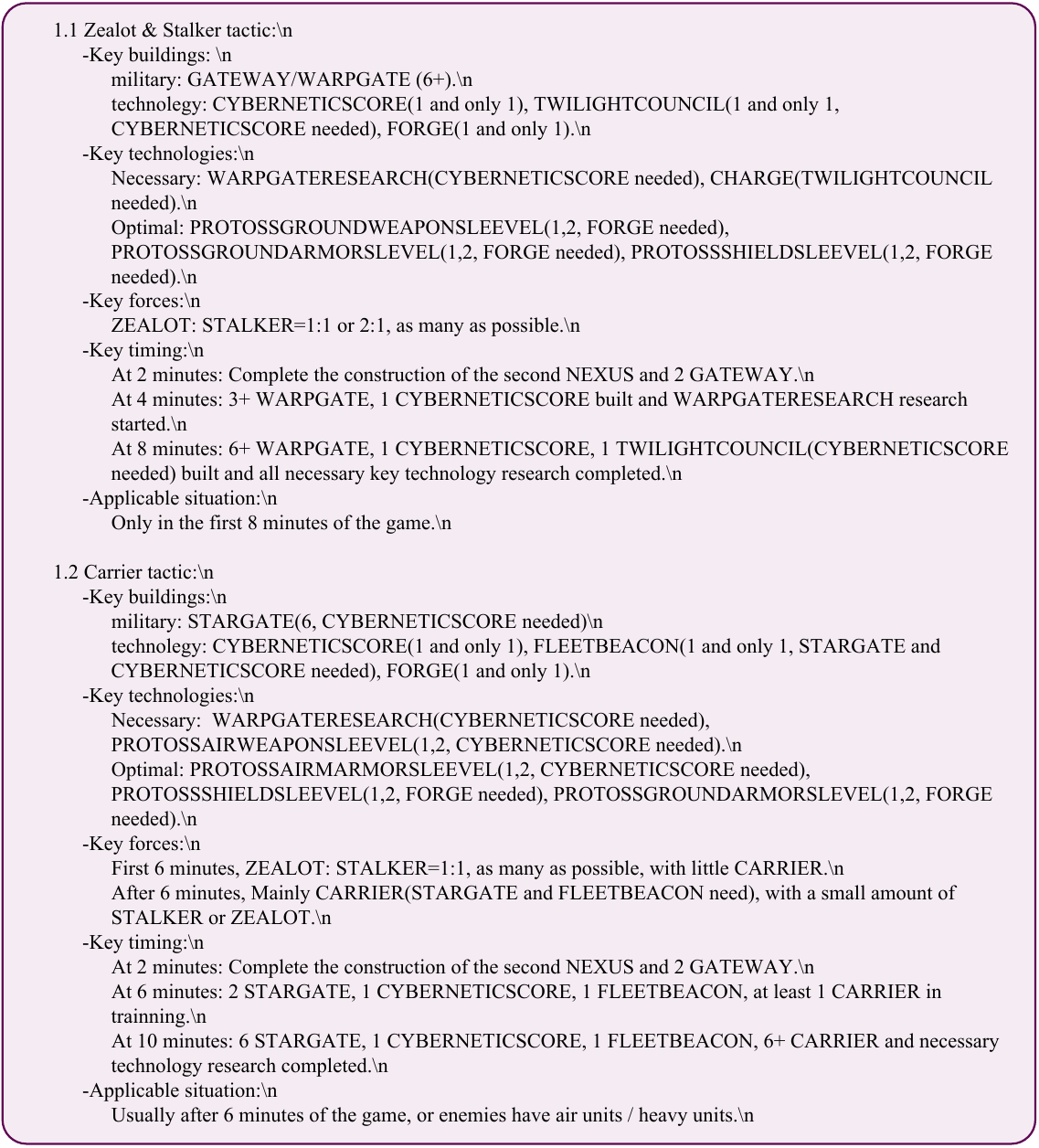}}
\caption{\textbf{Expert Tactic Prompt.} This part refers to \{Expert tactic\} part in Fig.A1. We experimentally constructed two tactics: 'Zealot \& Stalker tactic' and 'Carriers tactic', each tactic consists of five parts: Key building, Key technologies, Key forces, Key timing, and Applicable situation. Actually, it is possible to add more tactics.} 
\end{figure}

\begin{figure}
\centerline{\includegraphics[width=\textwidth]{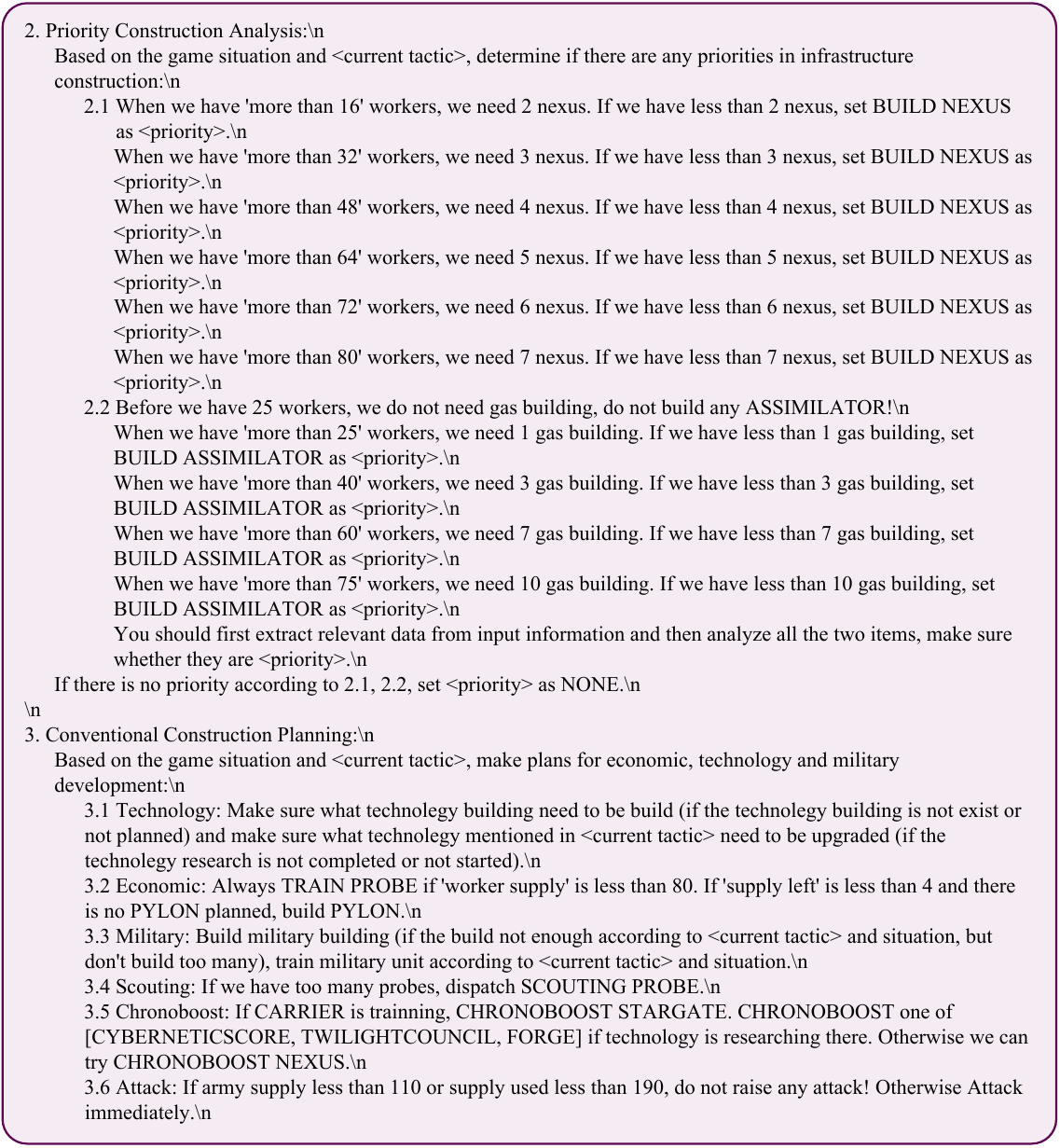}}
\caption{\textbf{Hierarchical Decision Prompt.} This prompt refers to 2. and 3. parts in Fig.A1. In the Priority Construction Analysis part, LLM should make sure whether there is <priority> and clearly describe it at the end of this part. In the Conventional Construction Planning part, LLM should make sure what to do in the aspects of Technology, Economy, Military,  Scouting, and Chronoboost and determine whether well developed to launch the final attack.} 
\end{figure}

\begin{figure}
\centerline{\includegraphics[width=\textwidth]{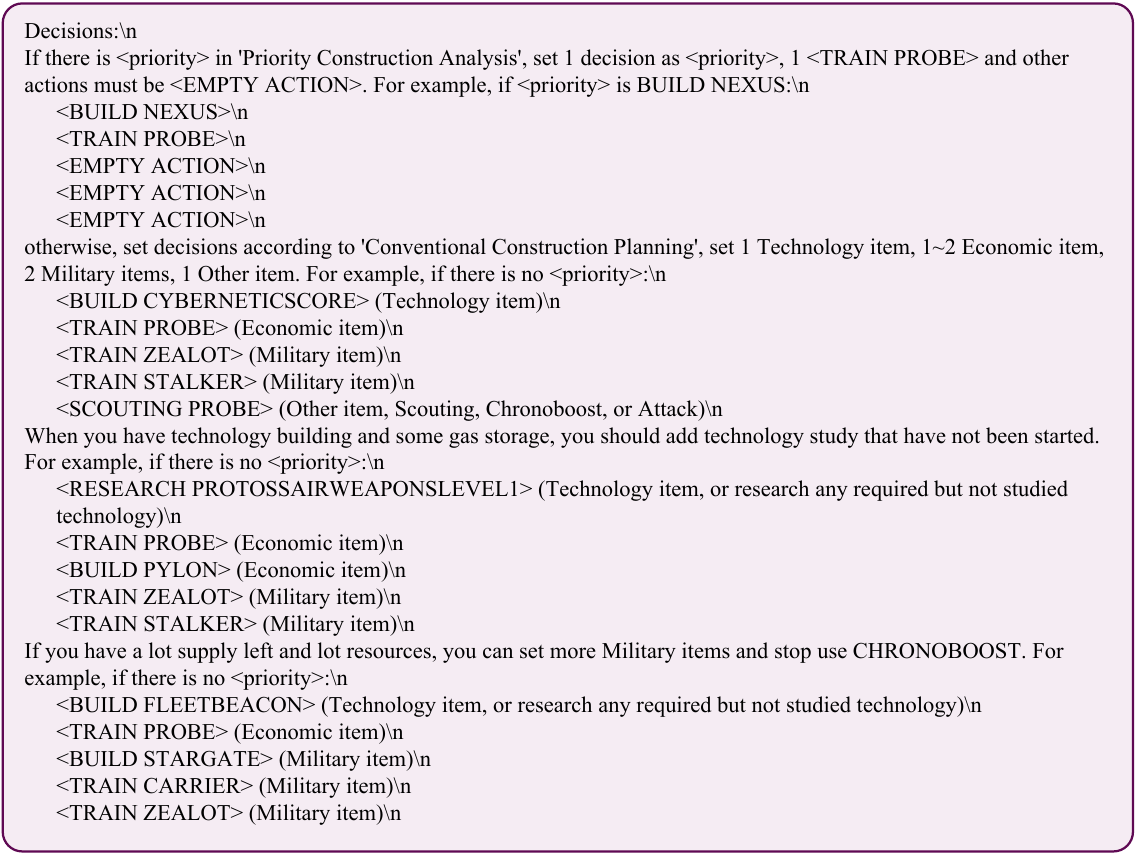}}
\caption{\textbf{Action Generating Prompt.} This part refers to 'Decisions' part in Fig.A1. In this part, we tell the LLM how to generate decisions in different situations.} 
\end{figure}

\begin{figure}
\centerline{\includegraphics[width=\textwidth]{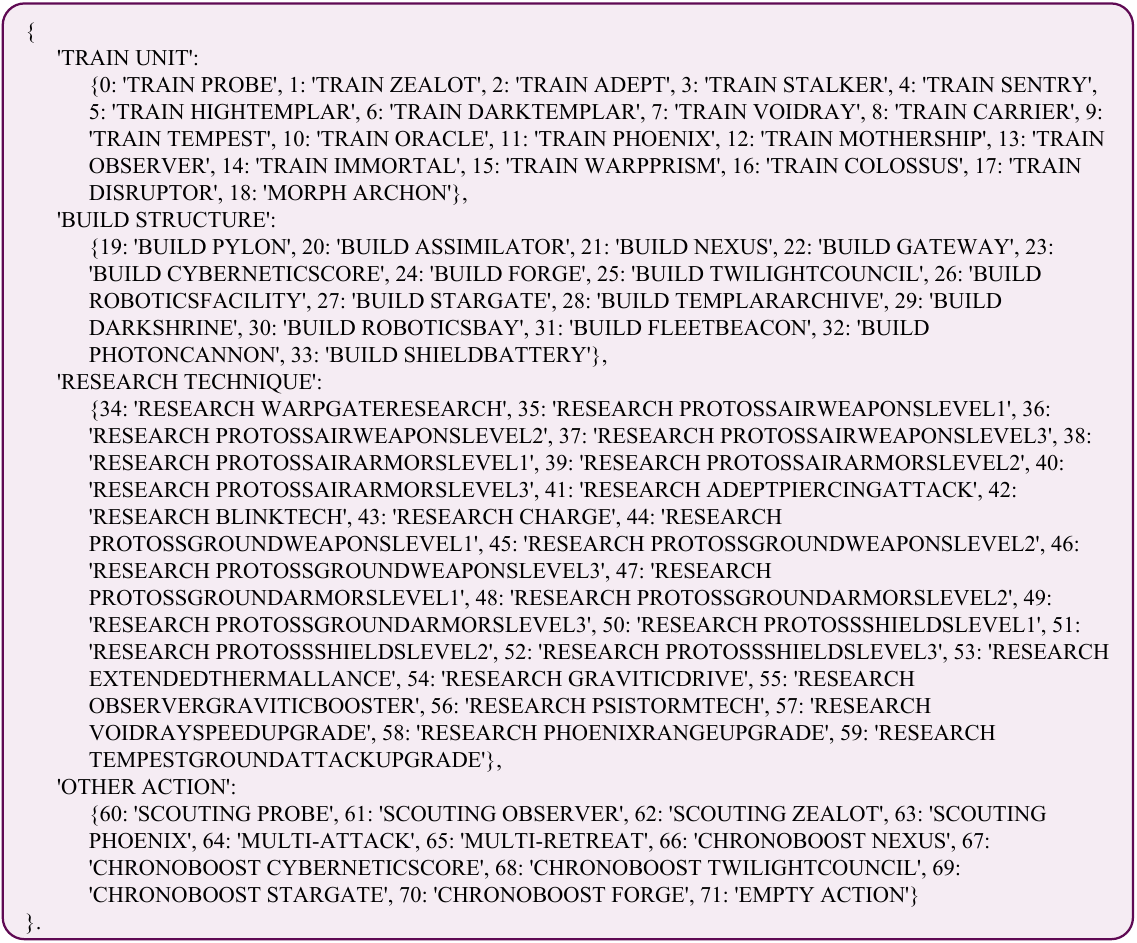}}
\caption{\textbf{Legal Action Library.} This part refers to \{Legal Action Library.\} part in Fig.A1. TextStarCraft II environment receives the text response of LLM, searching actions by locating segments shaped as '<TRAIN PROBE>' and discarding illegal actions. This part is necessary for generating legal actions.} 
\end{figure}

\clearpage
\textbf{A.1.2 Example input and output}

\begin{figure}
\centerline{\includegraphics[width=\textwidth]{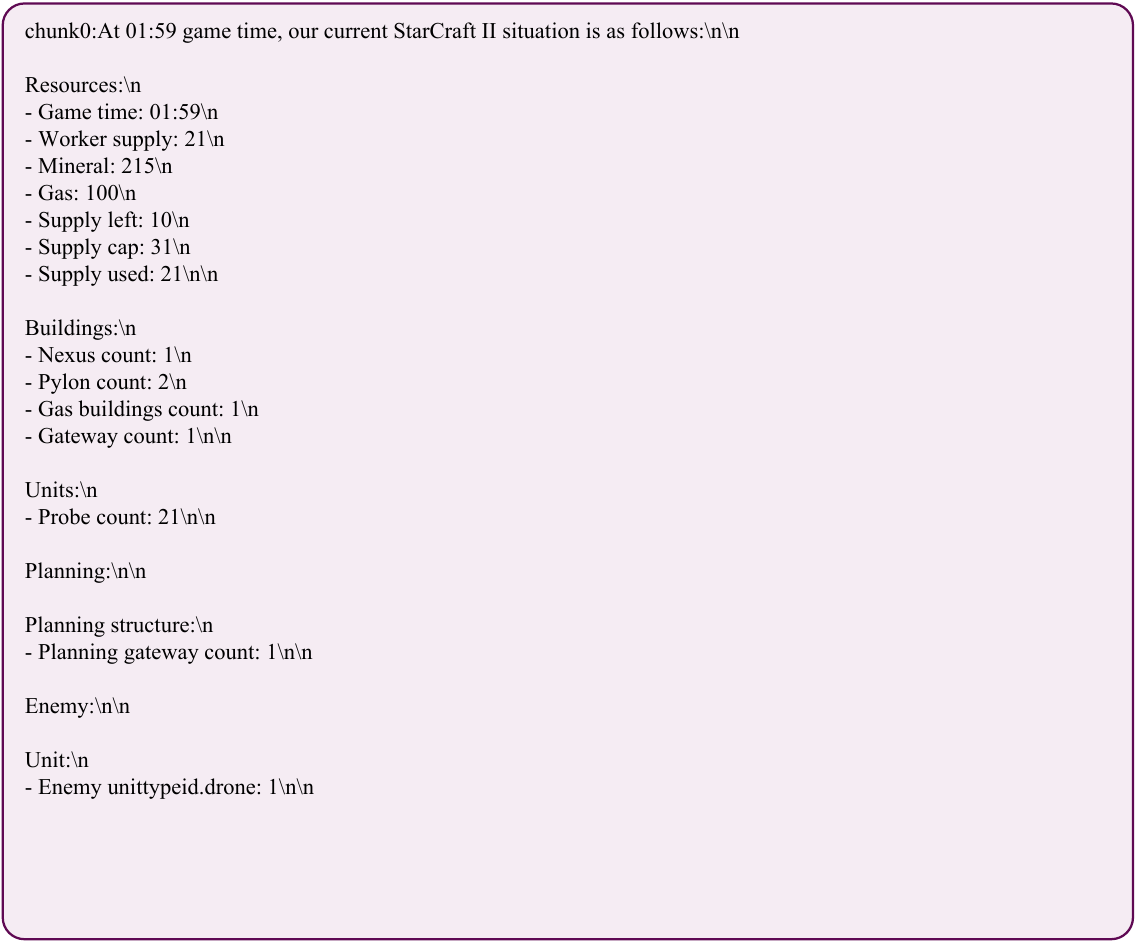}}
\caption{\textbf{Example Input Prompt.} This part tells LLM what the input observation looks like, and the example output prompt tells LLM how to analyze the game situation and make decisions.} 
\end{figure}

\begin{figure}
\centerline{\includegraphics[width=\textwidth]{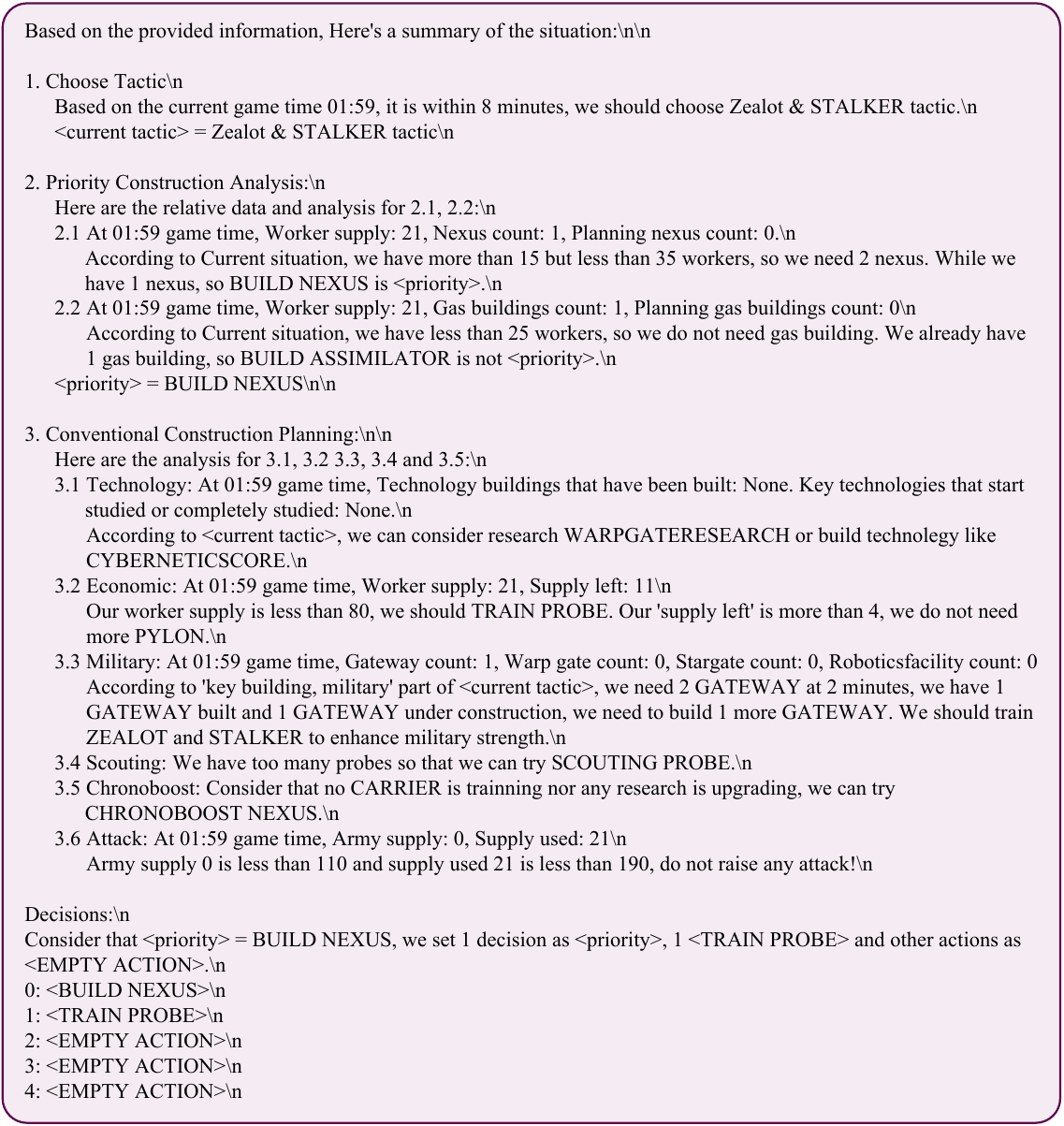}}
\caption{\textbf{Example Output Prompt.} This part tells LLM how to analyze game situations and make decisions. This part is essential because LLM will respond in a similar form as the example output prompt. We find that this part is highly related to the quality of reasoning and final decisions.} 
\end{figure}

\clearpage
\subsection*{A.2 Prompts in Ablation Study}
\textbf{A.2.1 Ablation Study 1: HEP without Expert Tactics}

\begin{figure}
\centerline{\includegraphics[width=\textwidth]{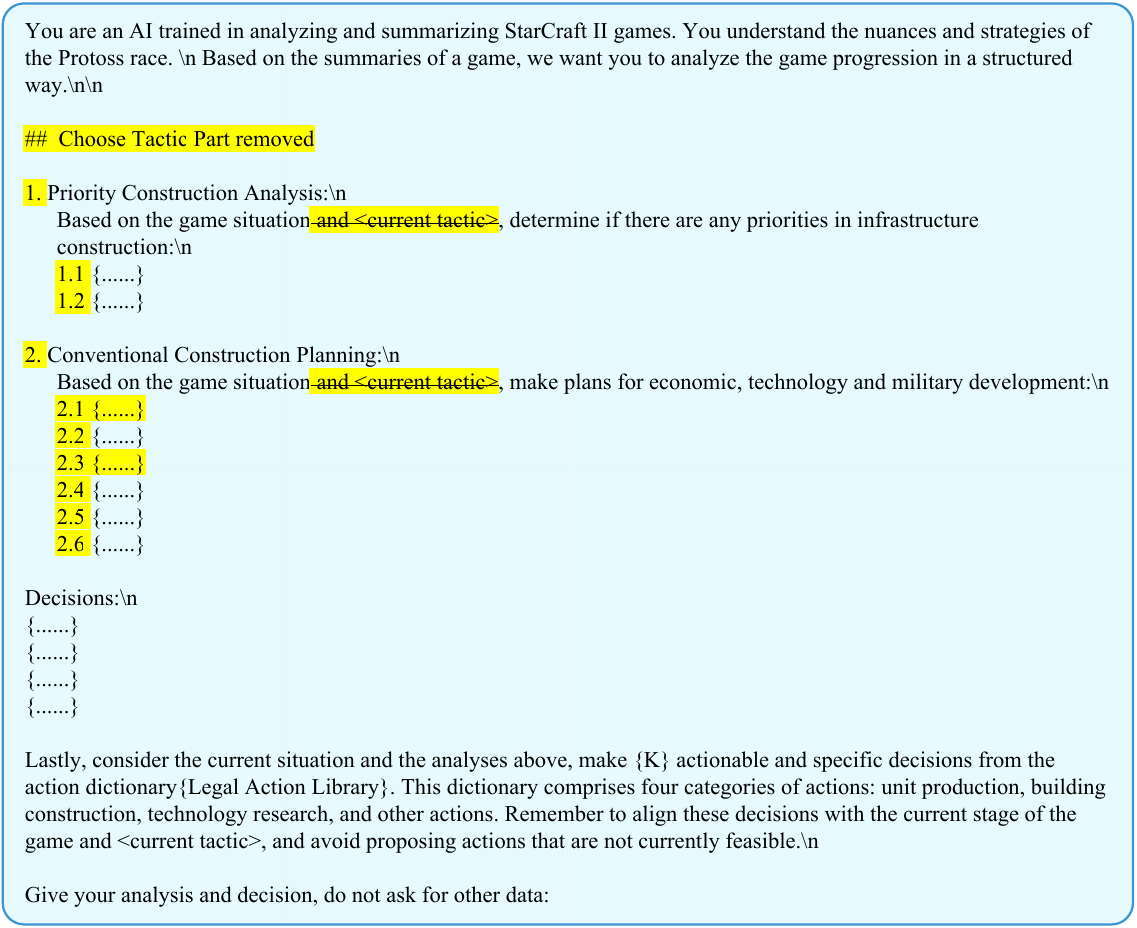}}
\caption{\textbf{General Structure of System Prompt in Ablation Study 1: HEP without Expert Tactics.} In Ablation Study 1, we remove the expert knowledge base and make corresponding changes to the areas where tactics are mentioned. Important modifications are highlighted.} 
\end{figure}

\begin{figure}
\centerline{\includegraphics[width=\textwidth]{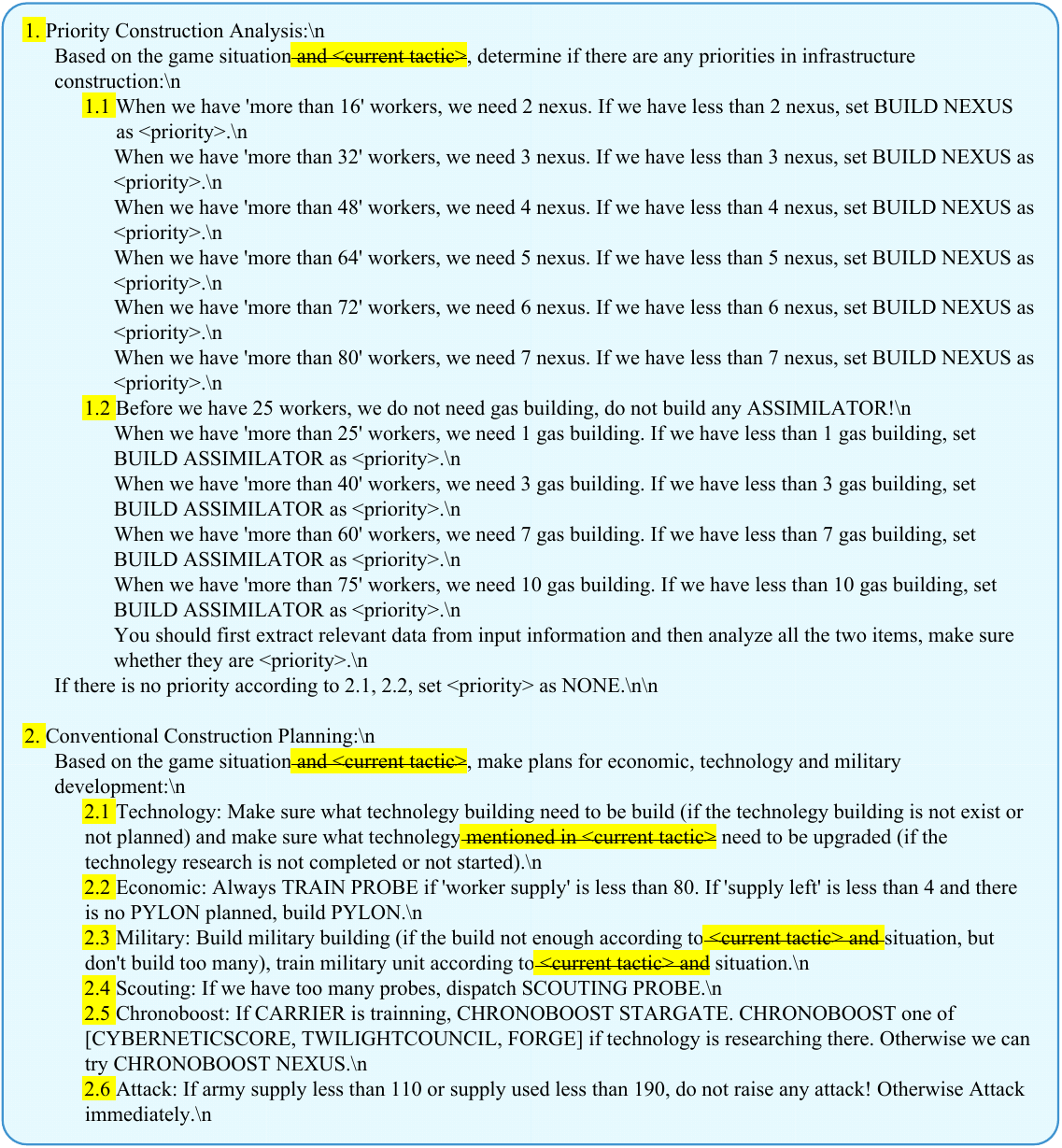}}
\caption{\textbf{Changed part in Ablation Study 1: HEP without Expert Tactics.} Considering that the 'Choose Tactic' part has been removed, we modify the indexing and remove segments where tactics are mentioned. Important modifications are highlighted.} 
\end{figure}

\begin{figure}
\centerline{\includegraphics[width=\textwidth]{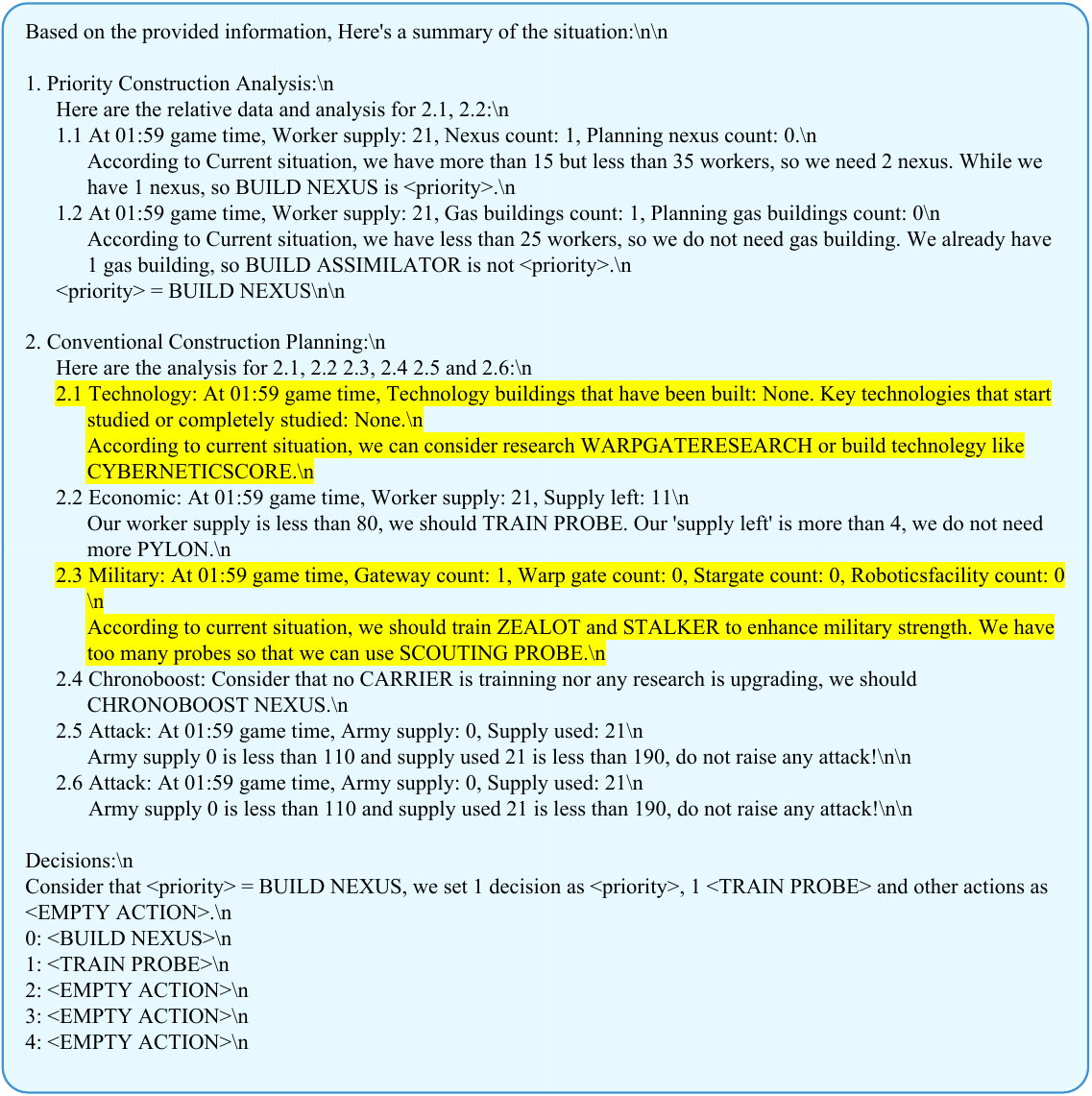}}
\caption{\textbf{Example output prompt in Ablation Study 1: HEP without Expert Tactics.} We remove segments where tactics are mentioned in the example output prompt. Important modifications are highlighted.} 
\end{figure}

\clearpage
\textbf{A.2.2 Ablation Study 2: HEP without Hierarchical Decision Prompt}
\begin{figure}
\centerline{\includegraphics[width=\textwidth]{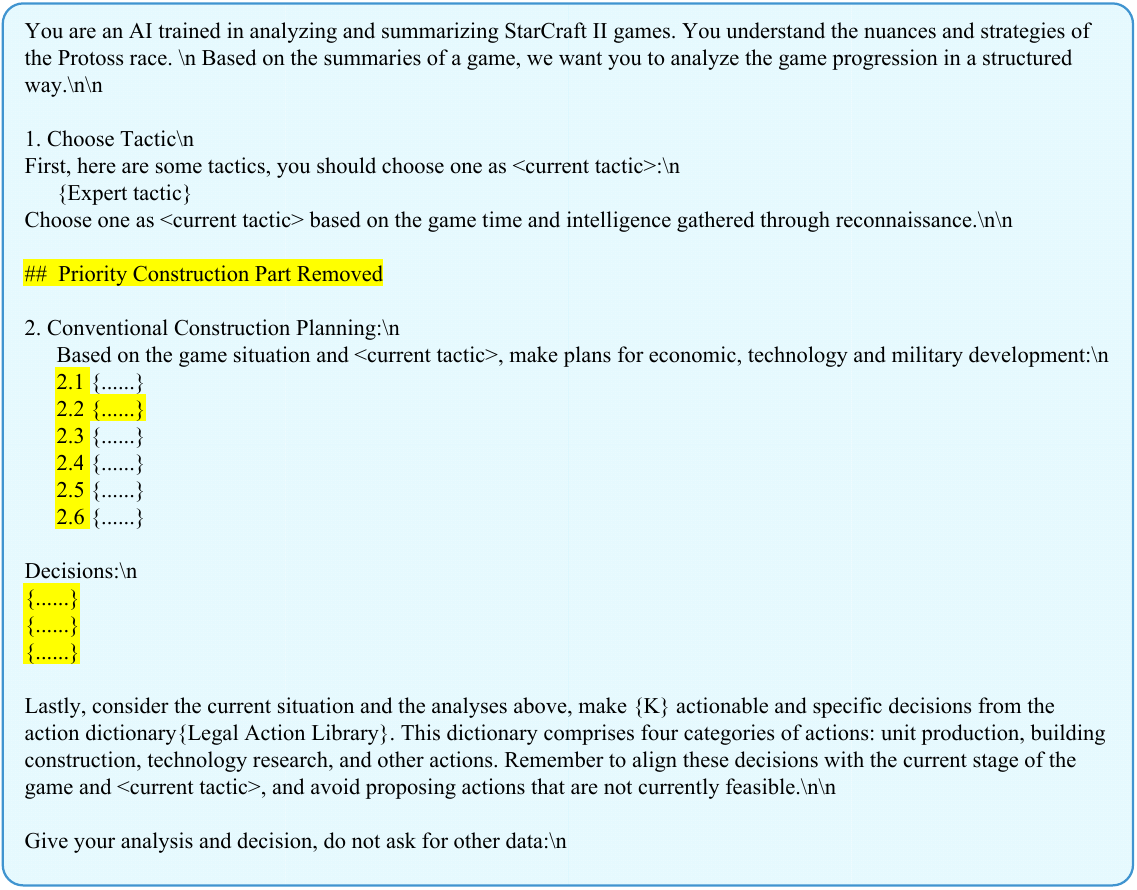}}
\caption{\textbf{General Structure of System Prompt in Ablation Study 2: HEP without Hierarchical Decision Prompt.} We remove the Priority Construction Part, but reserve the knowledge of when to build the Nexus and Gas building in 2.2. Corresponding modifications are made in other parts. Important modifications are highlighted.} 
\end{figure}

\begin{figure}
\centerline{\includegraphics[width=\textwidth]{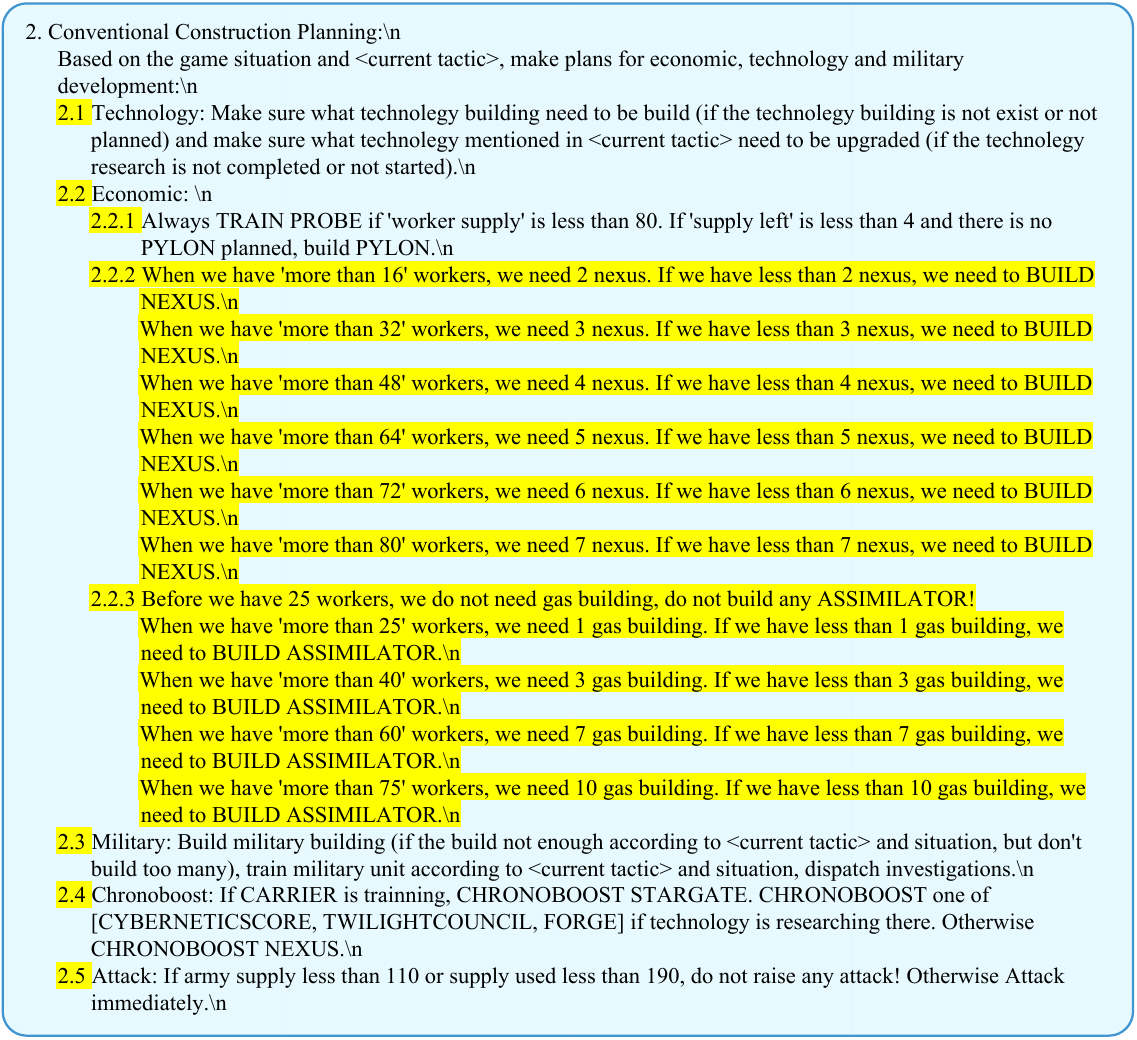}}
\caption{\textbf{Changed part (Conventional Construction) in Ablation Study 2: HEP without Hierarchical Decision Prompt.} In order to avoid the impact of lack of knowledge of when to build Nexus and Gas building, ensuring only the hierarchical decision structure affects the results, we rewrite this knowledge in 2.2. Important modifications are highlighted.} 
\end{figure}

\begin{figure}
\centerline{\includegraphics[width=\textwidth]{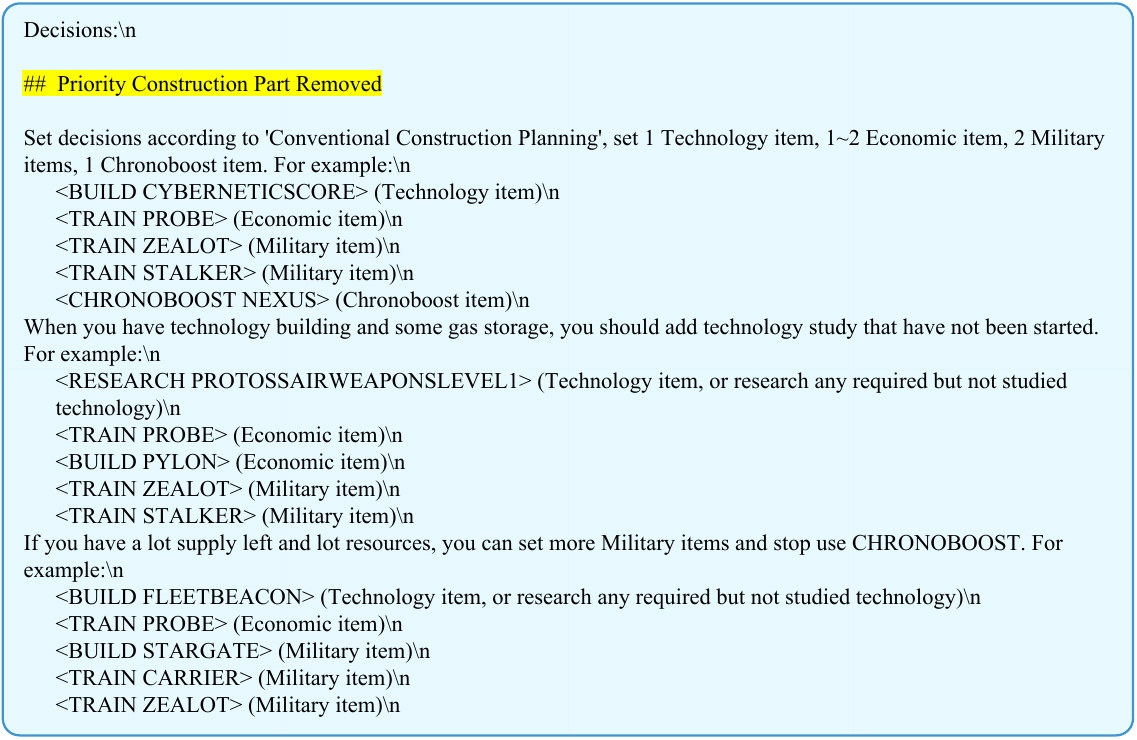}}
\caption{\textbf{Changed part (Decisions) in Ablation Study 2: HEP without Hierarchical Decision Prompt.} Considering that there is no <priority> in ablation study 2, we remove the segment that instructs LLM how to generate decisions when <priority> is not NONE. Important modifications are highlighted.} 
\end{figure}

\begin{figure}
\centerline{\includegraphics[width=\textwidth]{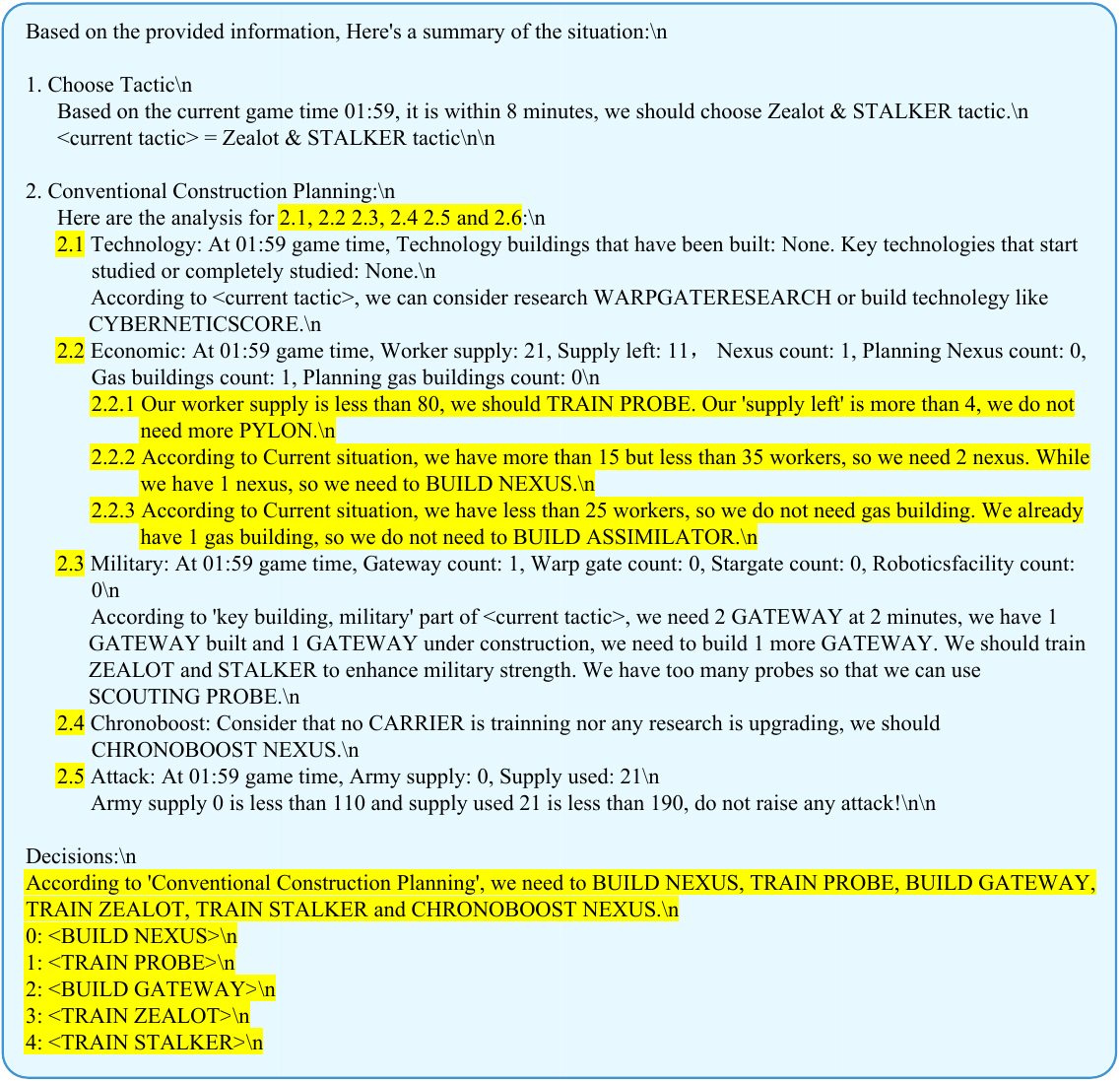}}
\caption{\textbf{Example output prompt in Ablation Study 2: HEP without Hierarchical Decision Prompt.} Relevant modifications are made in the example output prompt. Important modifications are highlighted.} 
\end{figure}




\clearpage
\section*{Appendix B. All Experiments Results}
\setcounter{figure}{0}
\renewcommand{\thefigure}{B\arabic{figure}}

We visualized more detailed data in Appendix B. From Fig.B1 to Fig.B9, we visualized the resources, supply and composition of the army, against Harder, VeryHard and Elite opponents. In addition, more detailed results of the ablation study can be found in Fig.B10 and Fig.B11.

\vspace{+2.4cm}

\begin{figure}
\centerline{\includegraphics[width=\textwidth]{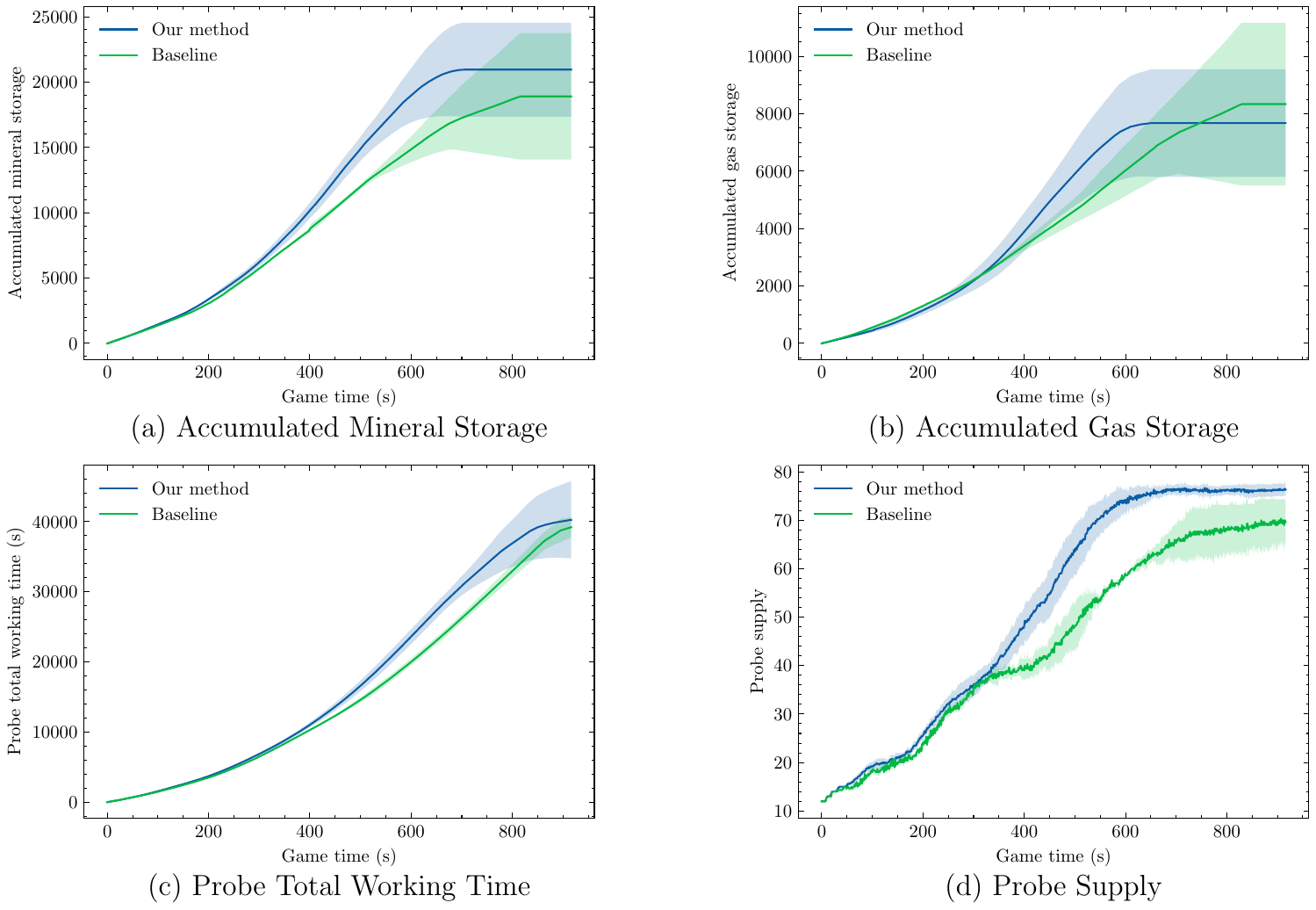}}
\caption{\textbf{Resources Data of Experiment against Hard built-in AI}} 
\end{figure}

\begin{figure}
\centerline{\includegraphics[width=\textwidth]{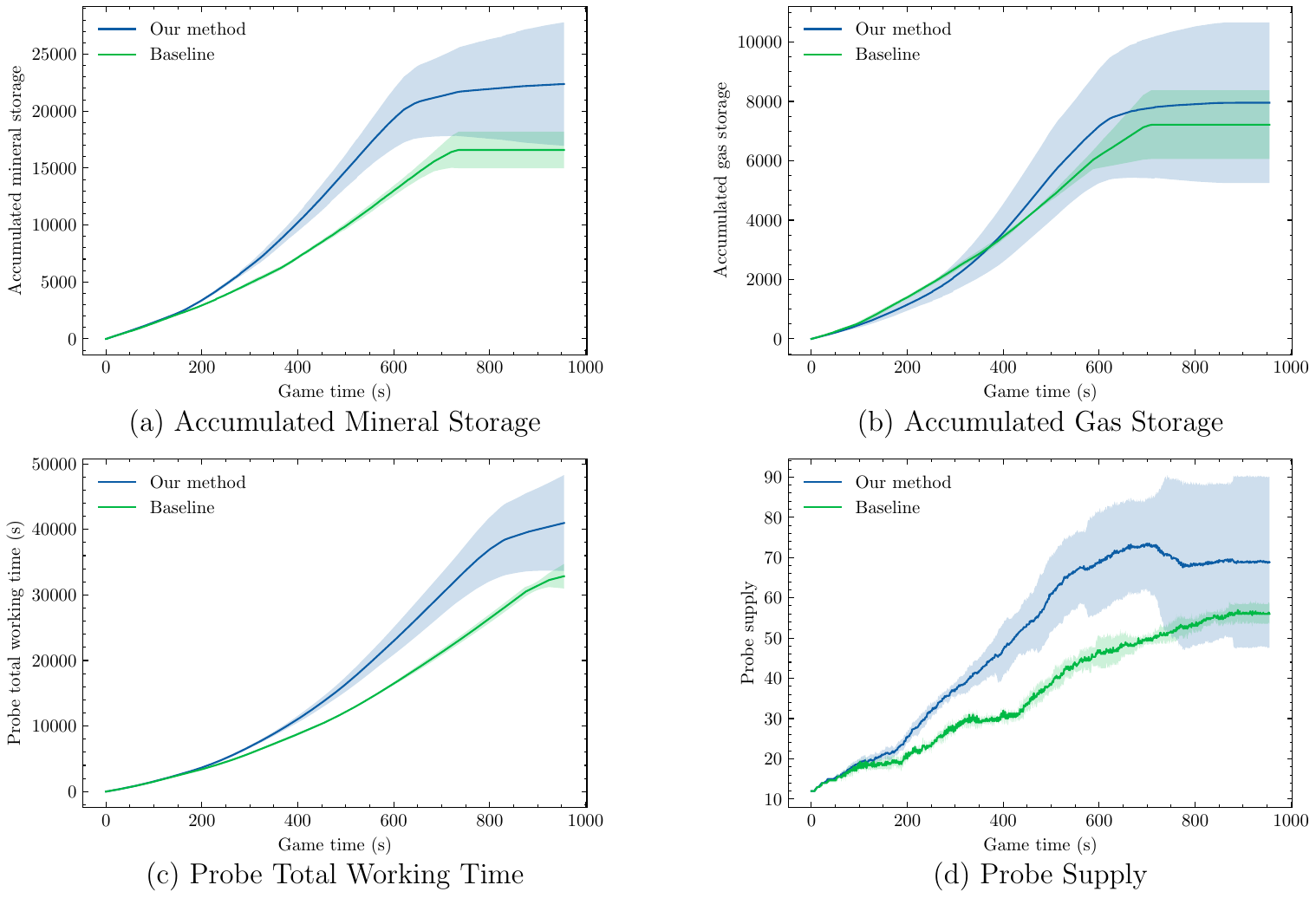}}
\caption{\textbf{Resources Data of Experiment against Harder built-in AI}} 
\end{figure}

\begin{figure}
\centerline{\includegraphics[width=\textwidth]{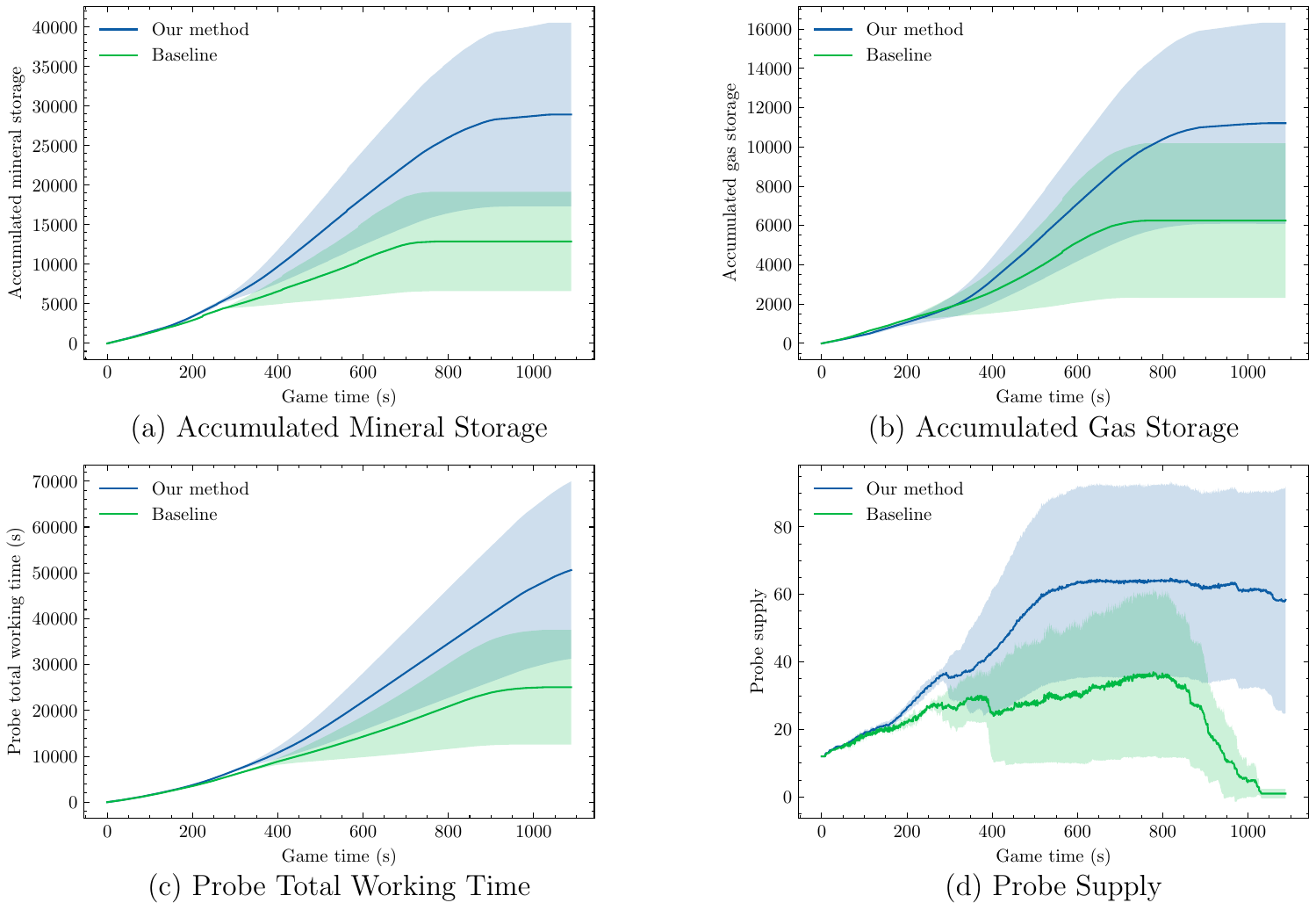}}
\caption{\textbf{Resources Data of Experiment against VeryHard built-in AI}} 
\end{figure}

\begin{figure}
\centerline{\includegraphics[width=\textwidth]{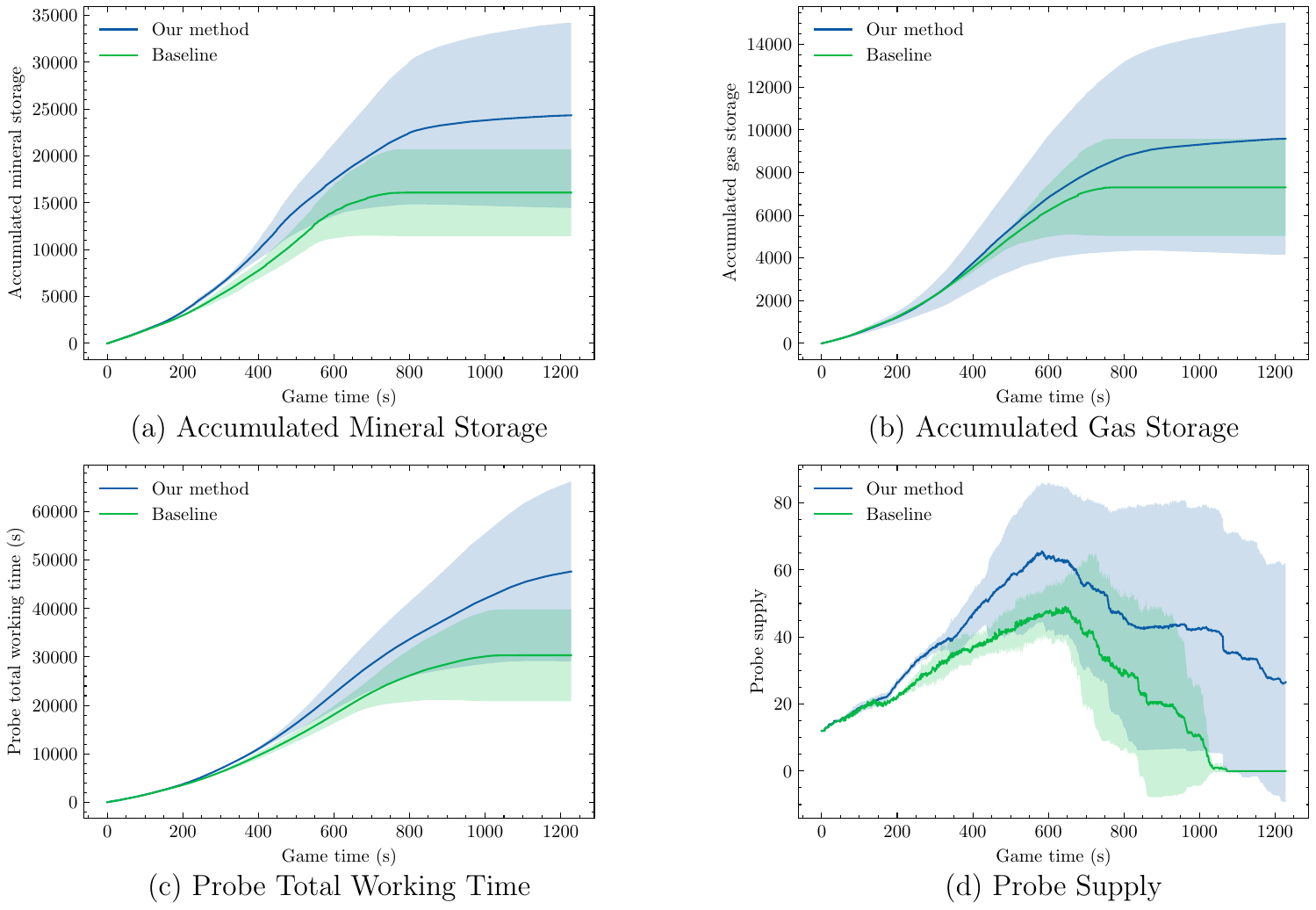}}
\caption{\textbf{Resources Data of Experiment against Elite built-in AI}} 
\end{figure}

\begin{figure}
\centerline{\includegraphics[width=\textwidth]{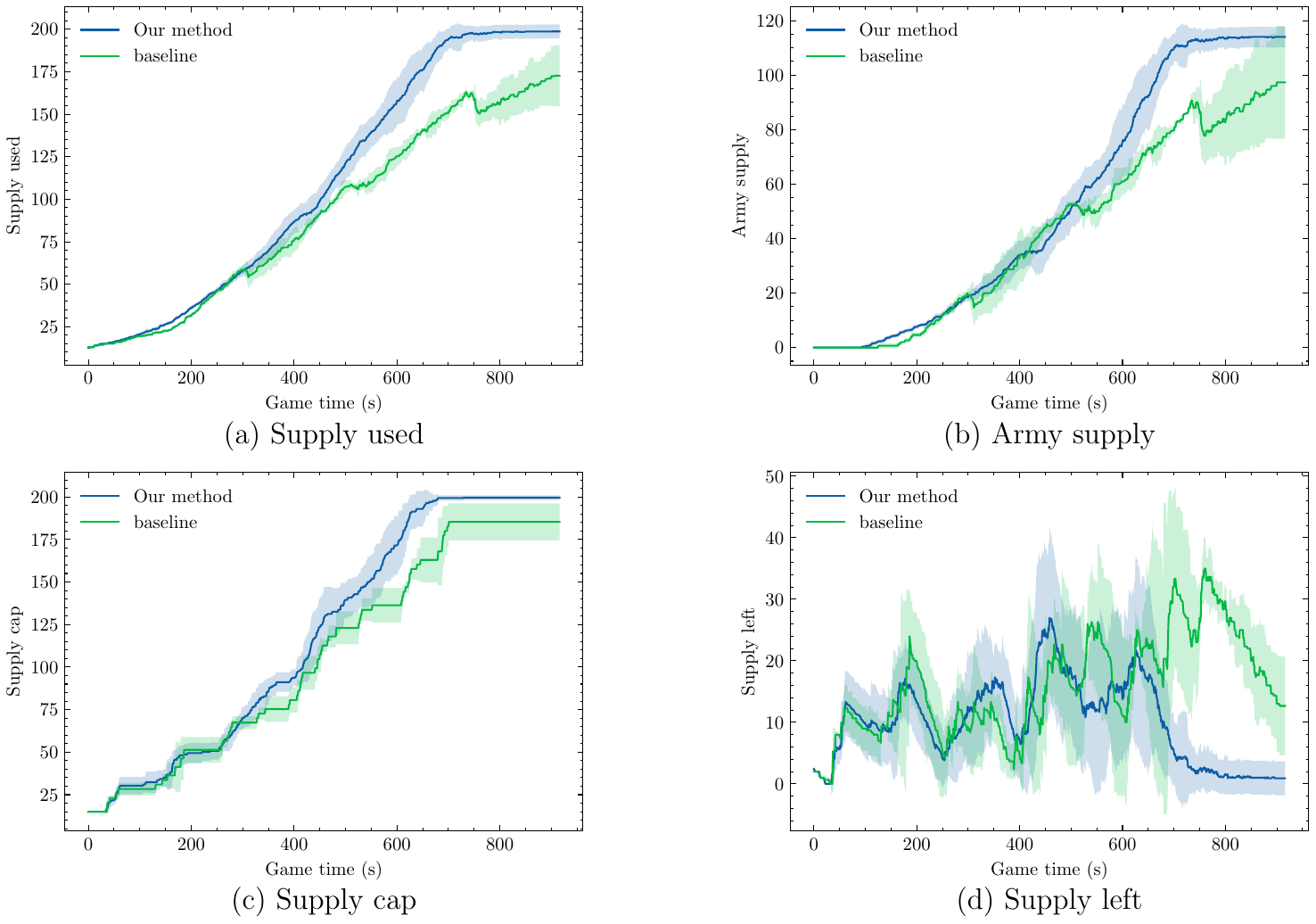}}
\caption{\textbf{Supply Data of Experiment against Hard built-in AI}} 
\end{figure}

\begin{figure}
\centerline{\includegraphics[width=\textwidth]{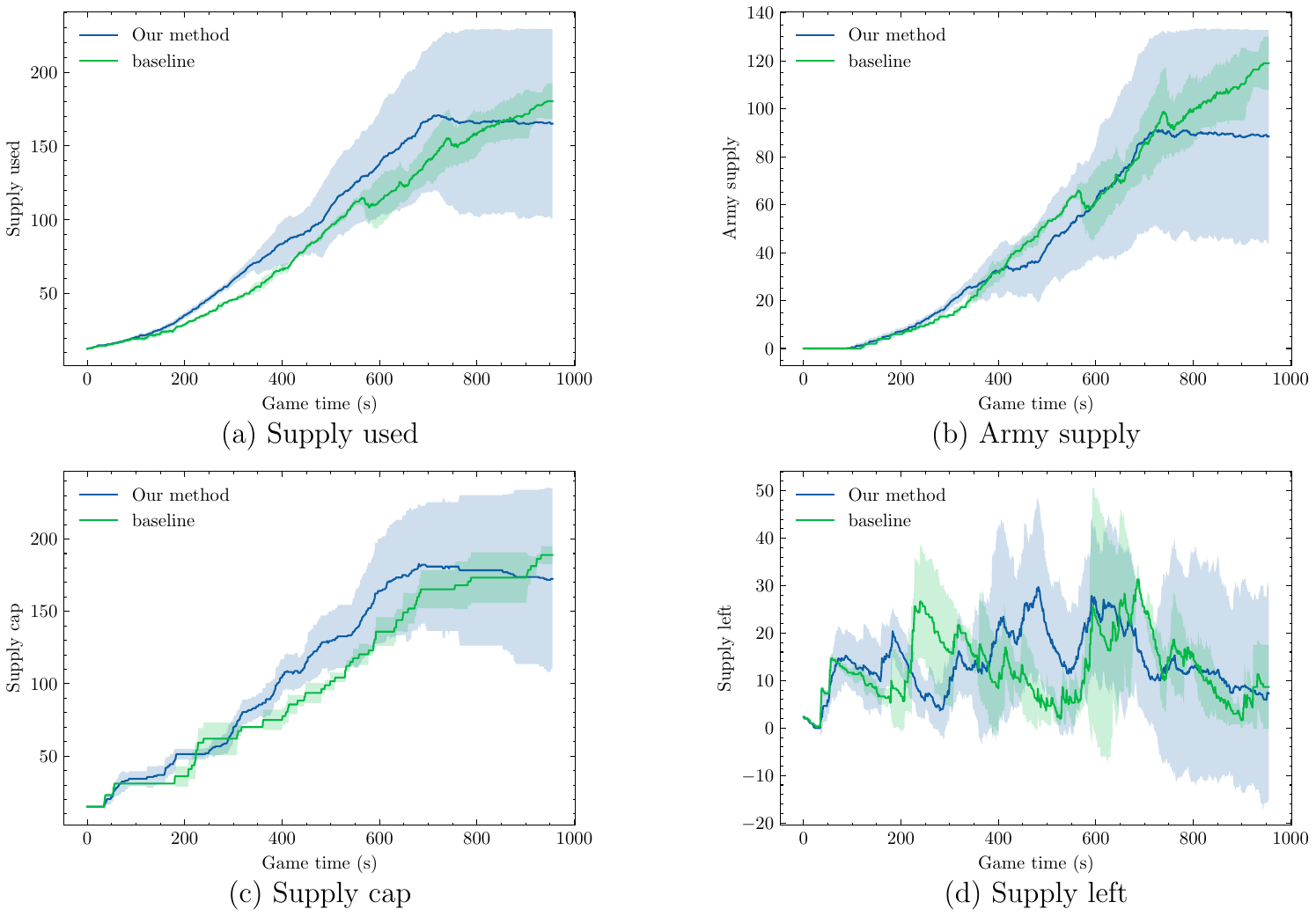}}
\caption{\textbf{Supply Data of Experiment against Harder built-in AI}} 
\end{figure}

\begin{figure}
\centerline{\includegraphics[width=\textwidth]{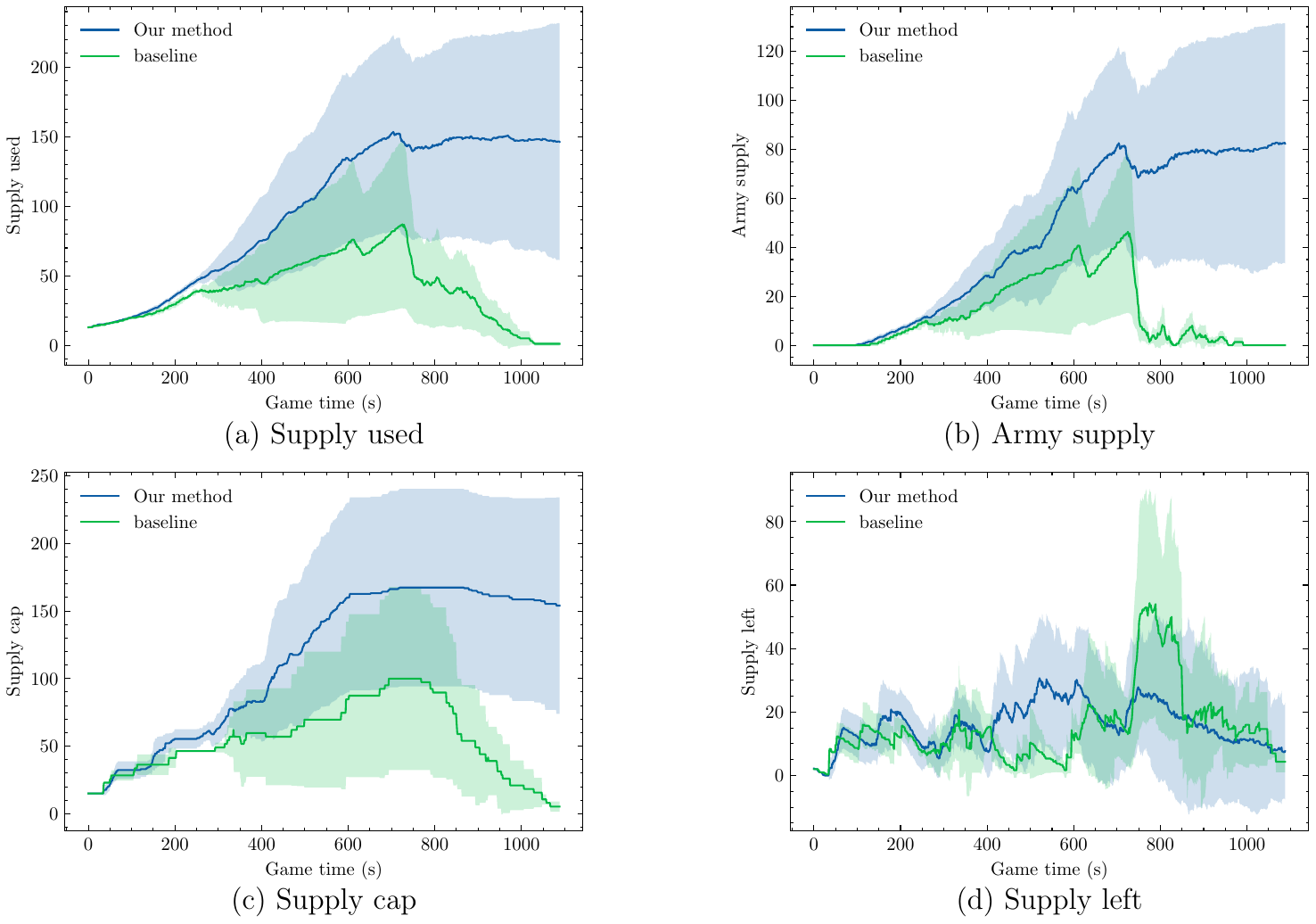}}
\caption{\textbf{Supply Data of Experiment against VeryHard built-in AI}} 
\end{figure}

\begin{figure}
\centerline{\includegraphics[width=\textwidth]{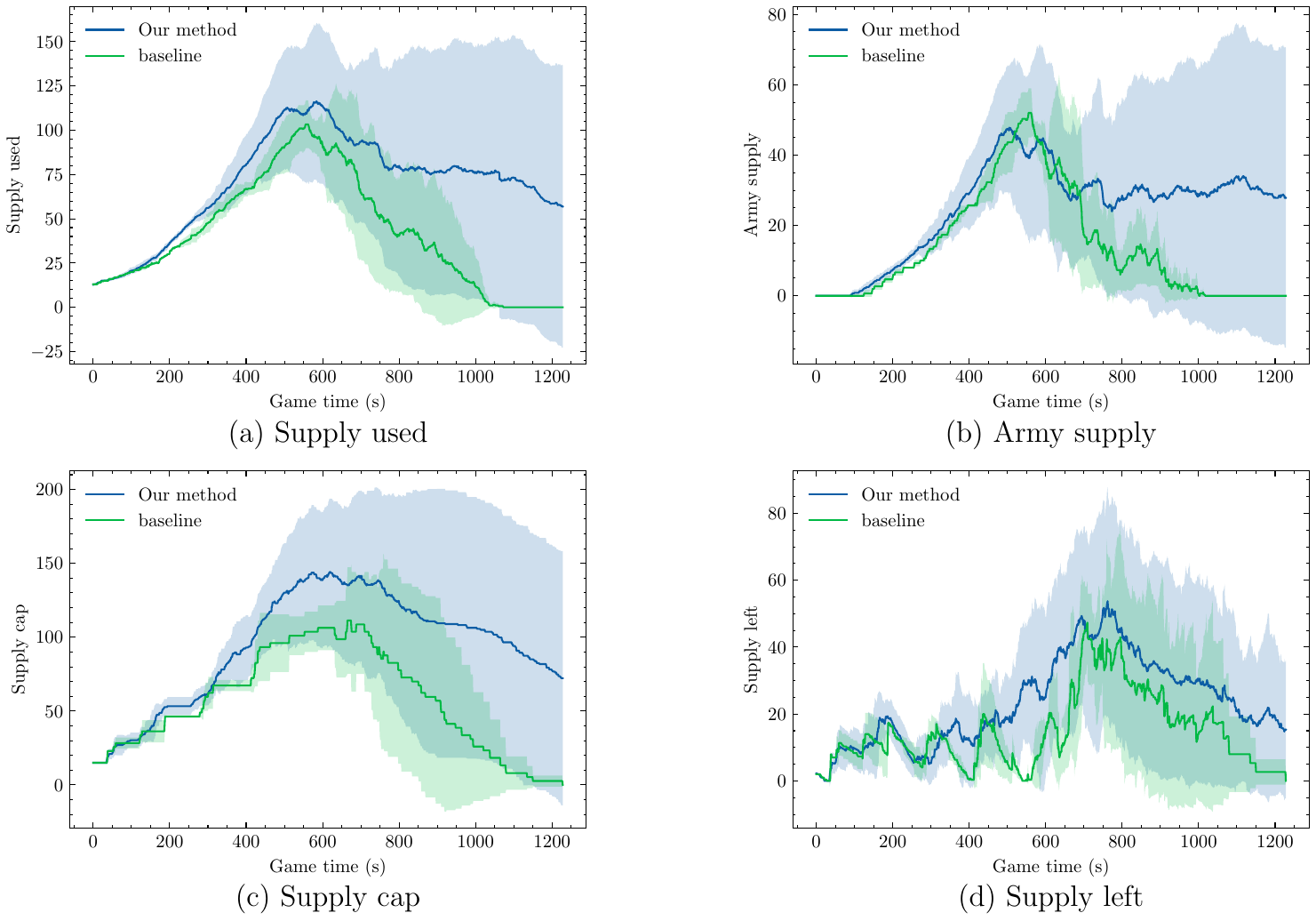}}
\caption{\textbf{Supply Data of Experiment against Elite built-in AI}} 
\end{figure}

\begin{figure}
\centerline{\includegraphics[width=\textwidth]{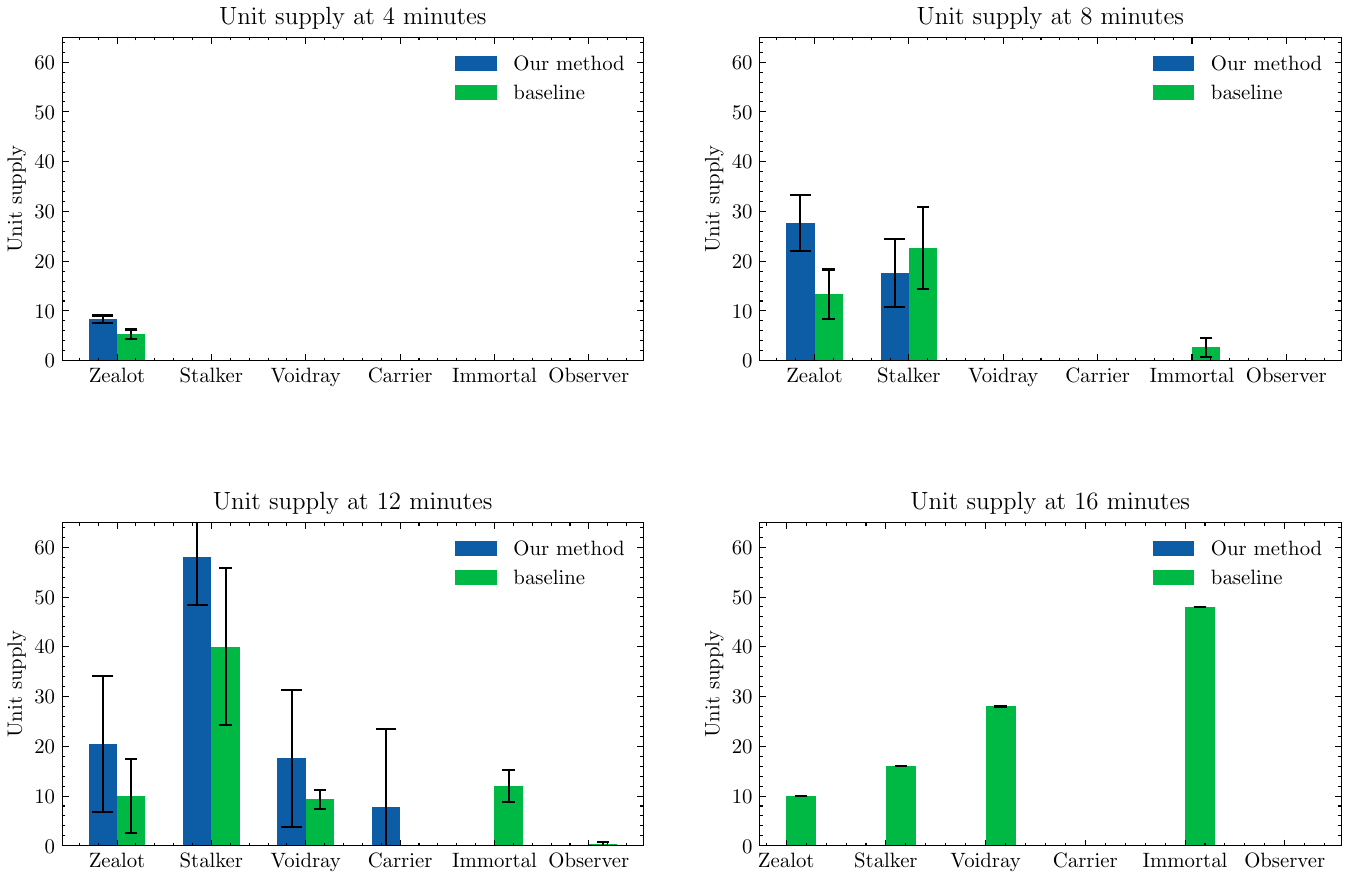}}
\caption{\textbf{Composition of the army of Experiment against Hard built-in AI}} 
\end{figure}

\begin{figure}
\centerline{\includegraphics[width=\textwidth]{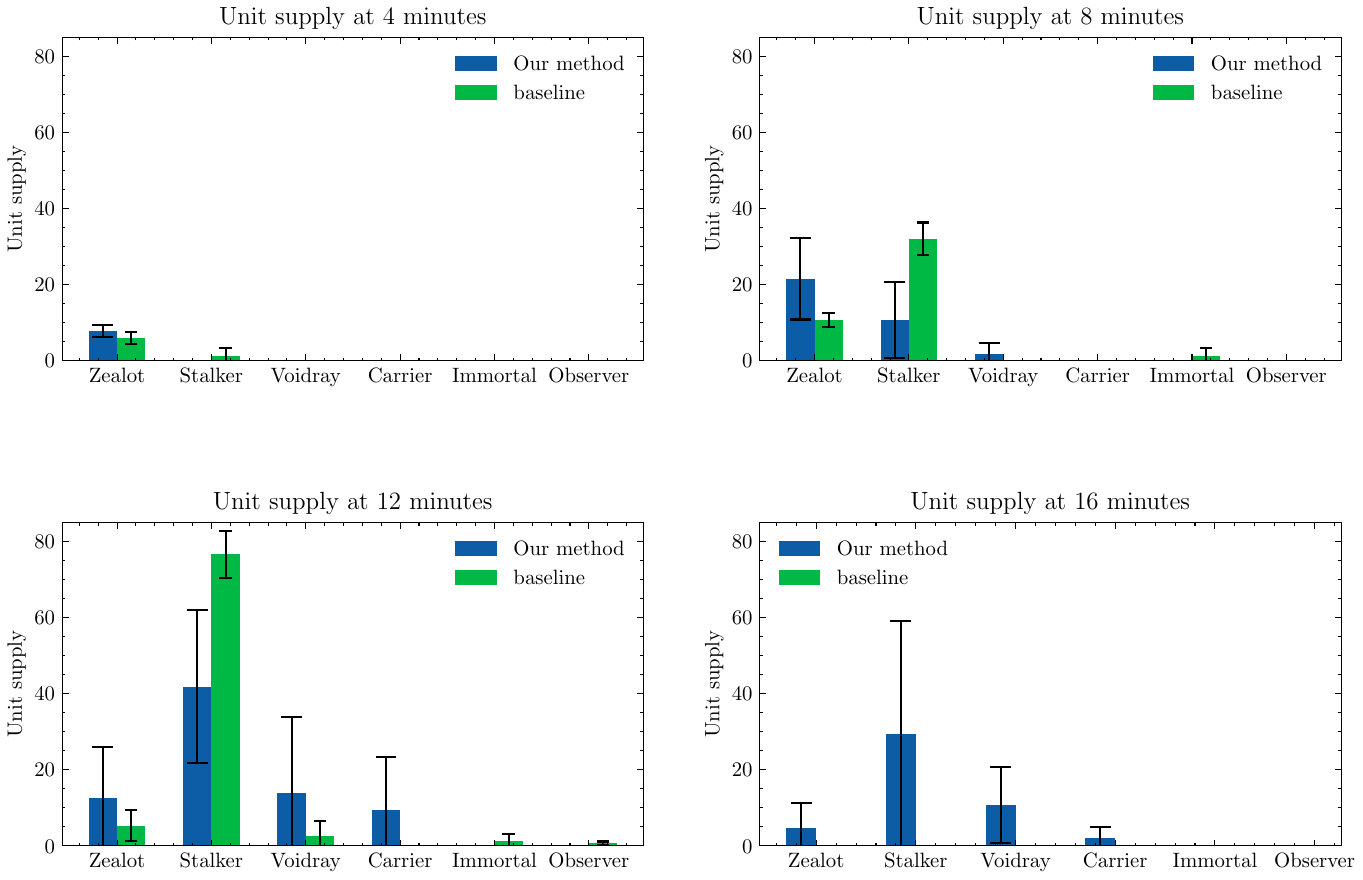}}
\caption{\textbf{Composition of the army of Experiment against Harder built-in AI}} 
\end{figure}

\begin{figure}
\centerline{\includegraphics[width=\textwidth]{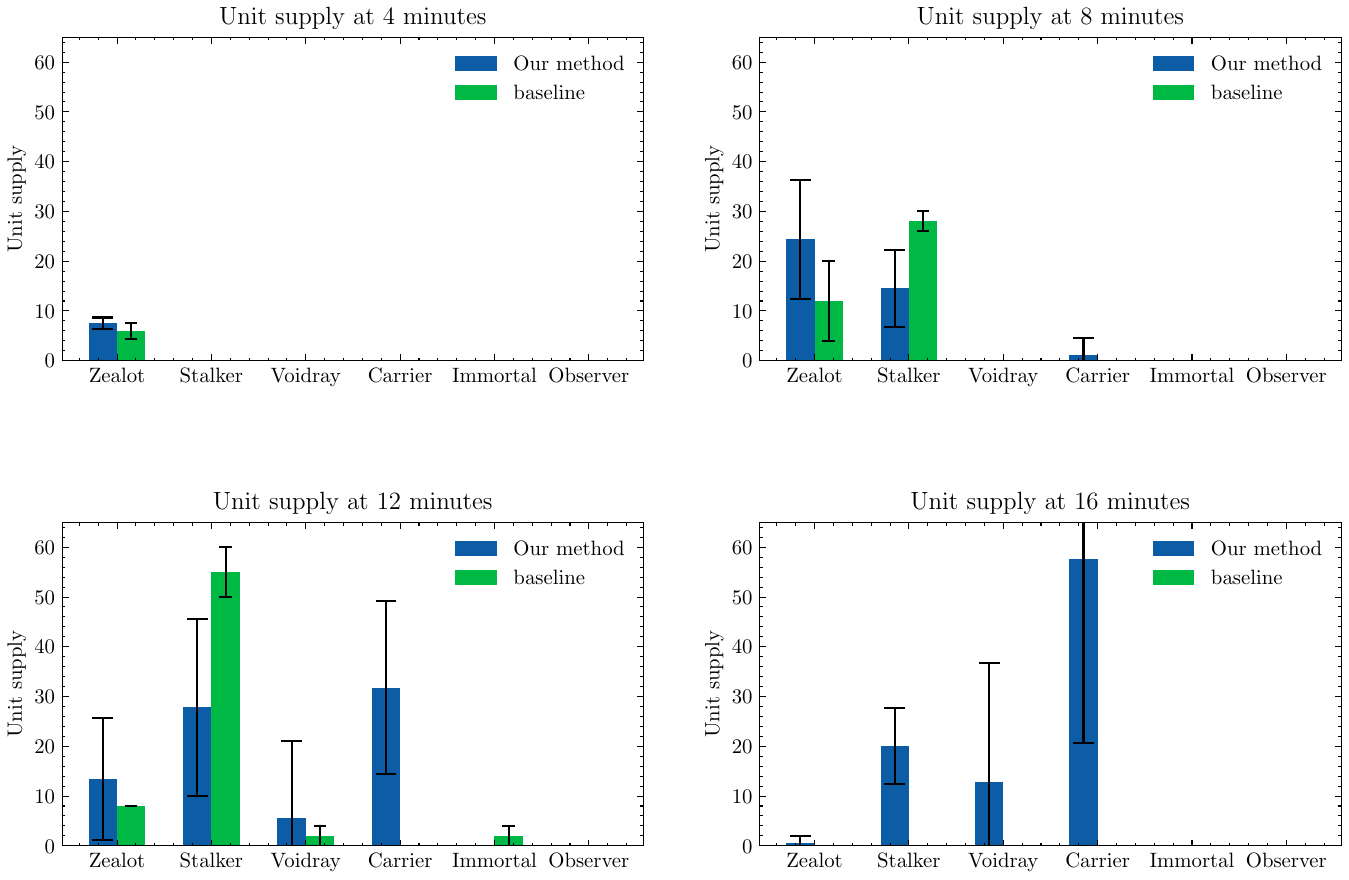}}
\caption{\textbf{Composition of the army of Experiment against VeryHard built-in AI}} 
\end{figure}

\begin{figure}
\centerline{\includegraphics[width=\textwidth]{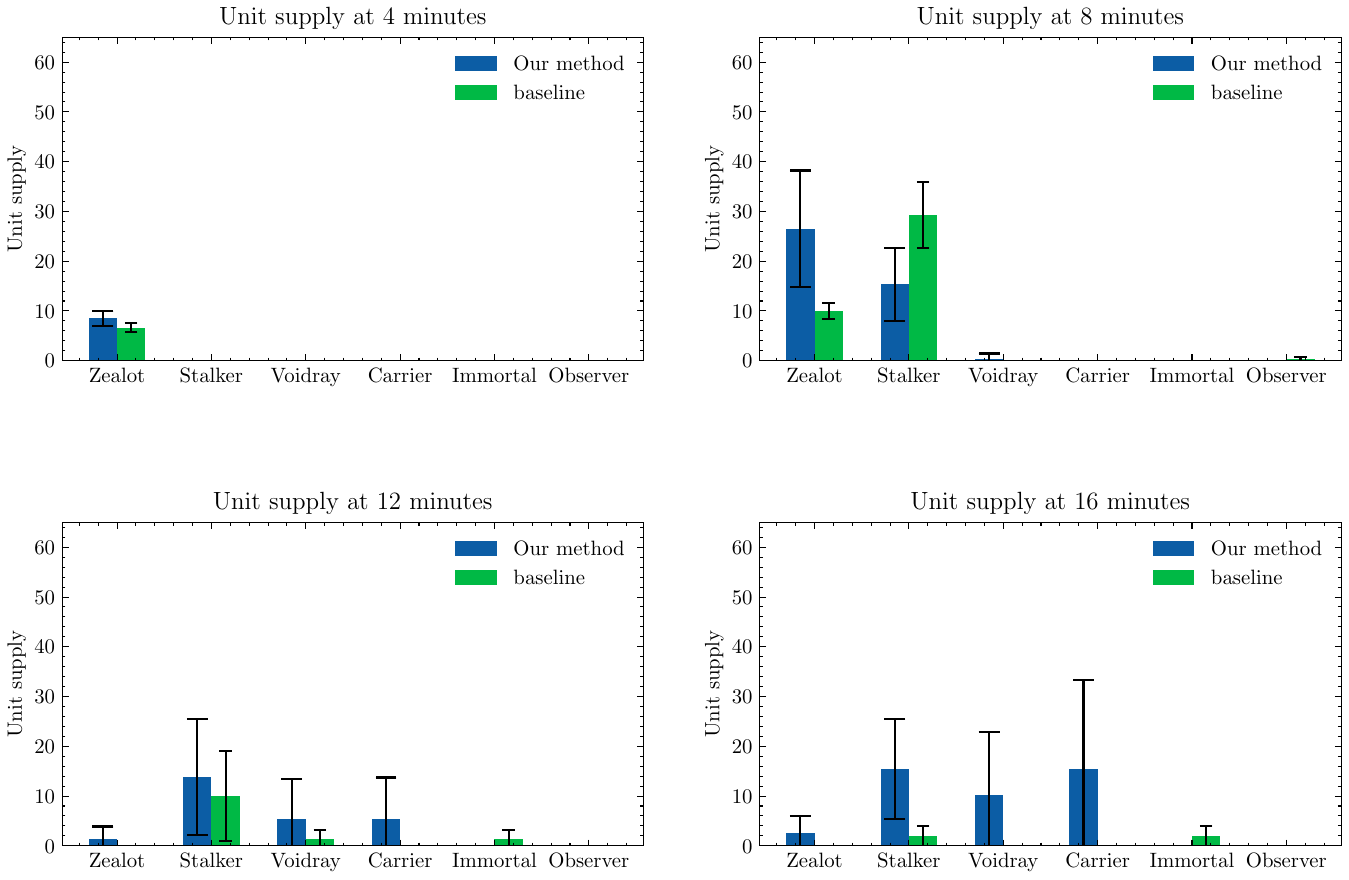}}
\caption{\textbf{Composition of the army of Experiment against Elite built-in AI}} 
\end{figure}

\begin{figure}
\centerline{\includegraphics[width=\textwidth]{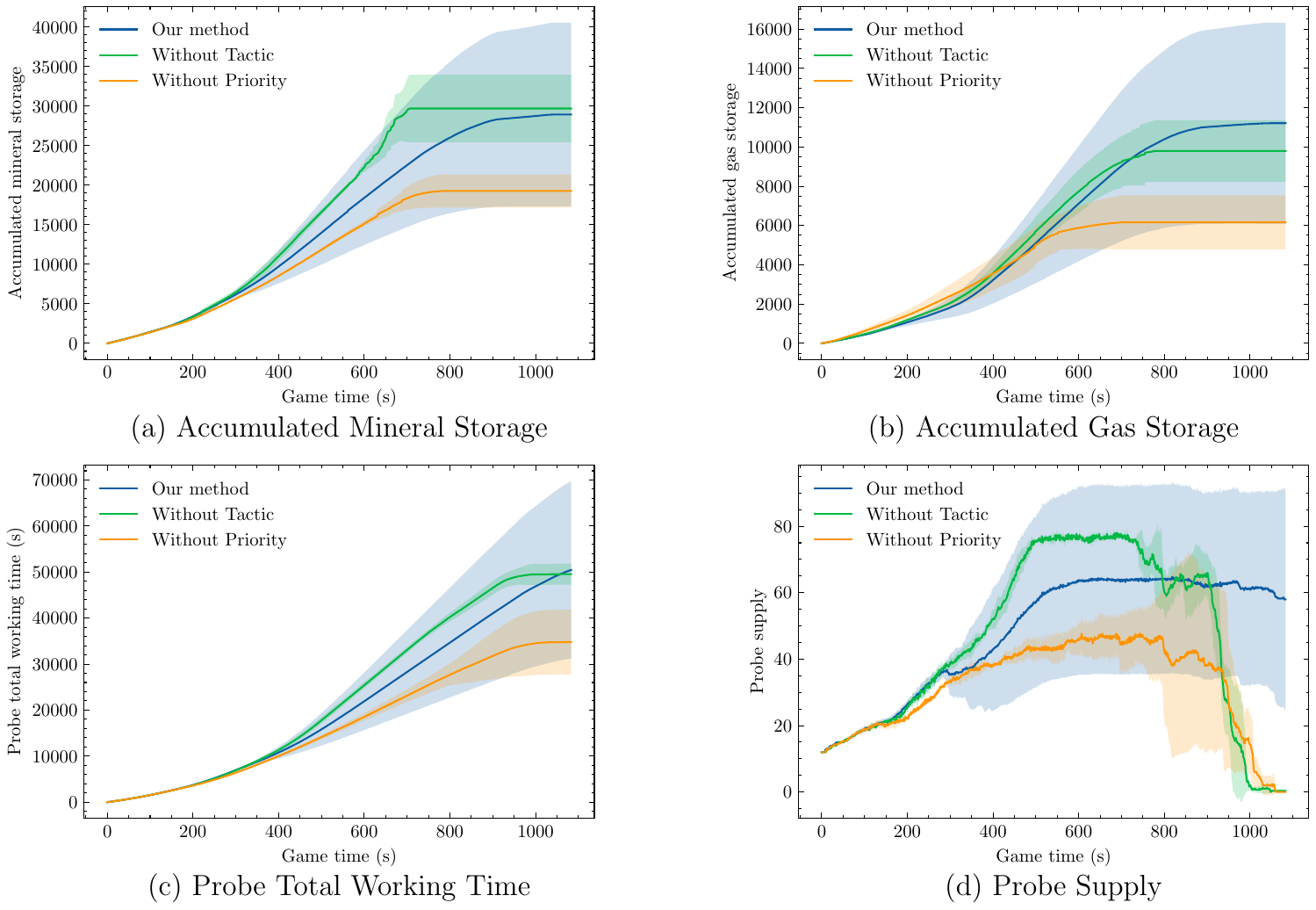}}
\caption{\textbf{Detailed Resources Data of Ablation Experiment}} 
\end{figure}

\begin{figure}
\centerline{\includegraphics[width=\textwidth]{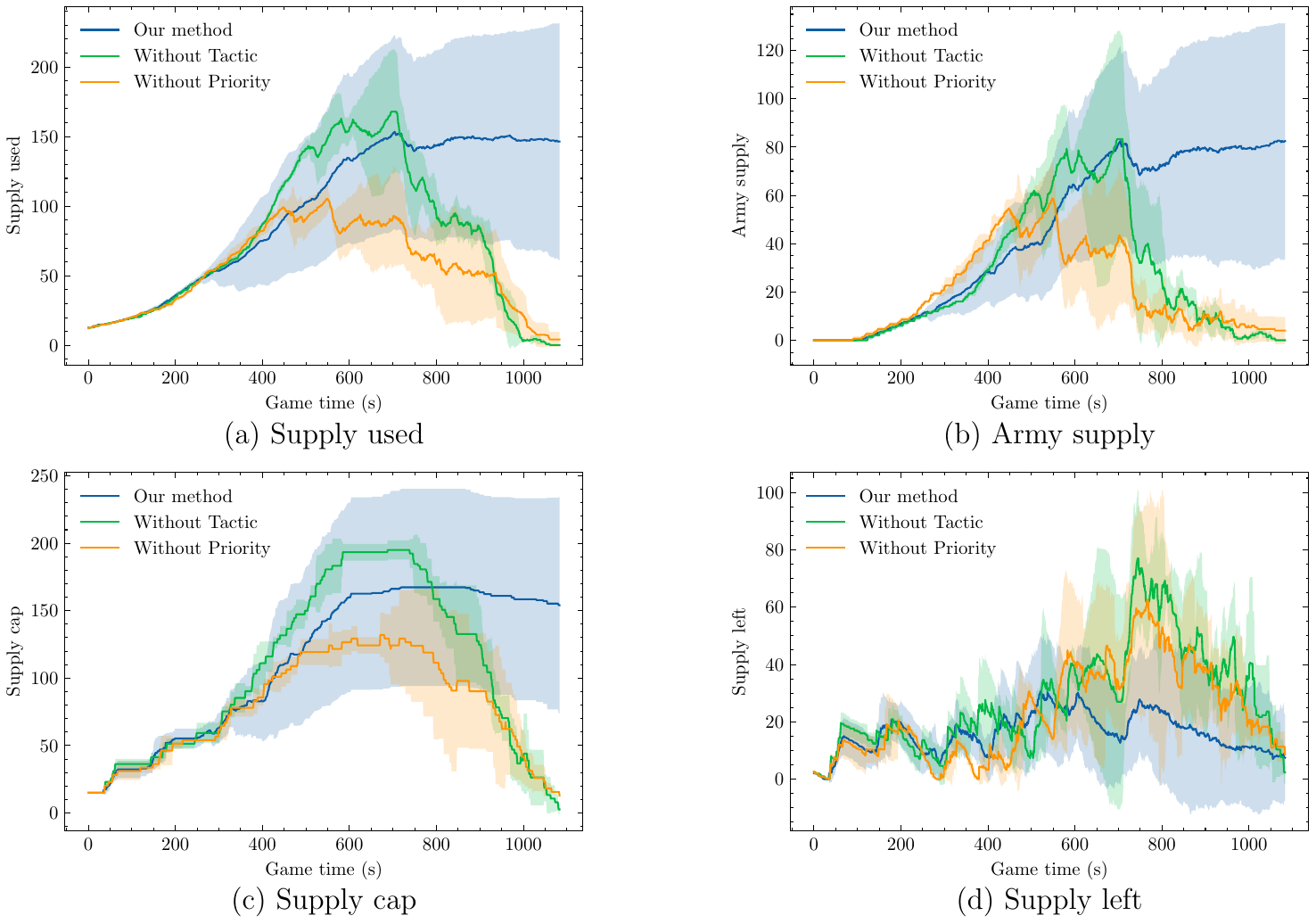}}
\caption{\textbf{Detailed Supply Data of Ablation Experiment}} 
\end{figure}

\clearpage
\section*{Apendix C. Suggetions in Prompt Engineering}
\setcounter{figure}{0}
\renewcommand{\thefigure}{C\arabic{figure}}
\subsection*{C.1 Describe Game Time to Better Extract Observed Data}

In the example output, we describe the game time before extracting the game data. Considering that the game time in text observation is usually different from the time in the example input, describing the game time before extracting data helps the agent find the correct data.

\begin{figure}
\centerline{\includegraphics[width=\textwidth]{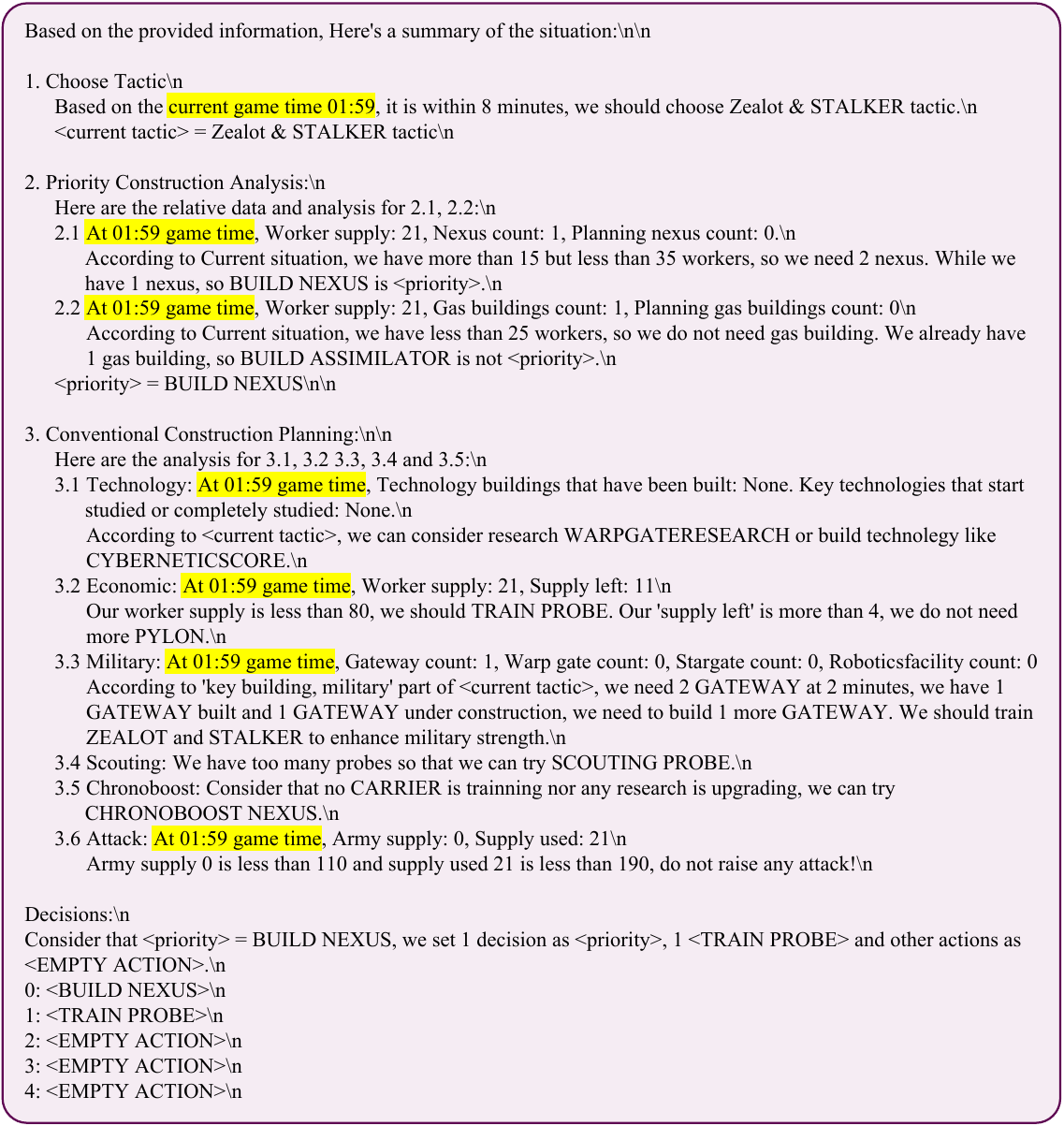}}
\caption{\textbf{Time Description in Example output} Whenever we need game data, we describe the game time, helping the agent better locating where the data is in the 5000-tokens-long text.}
\end{figure}

\clearpage
\subsection*{C.2 Always Extract Relevant Data Before Analyzing}

We find that extracting relevant data before analyzing is important to increase the accuracy of the analysis. Without data extraction, LLM-hallucination appears more frequently, leading to errors such as direct copying of the analysis and conclusion from example output.

\begin{figure}
\centerline{\includegraphics[width=\textwidth]{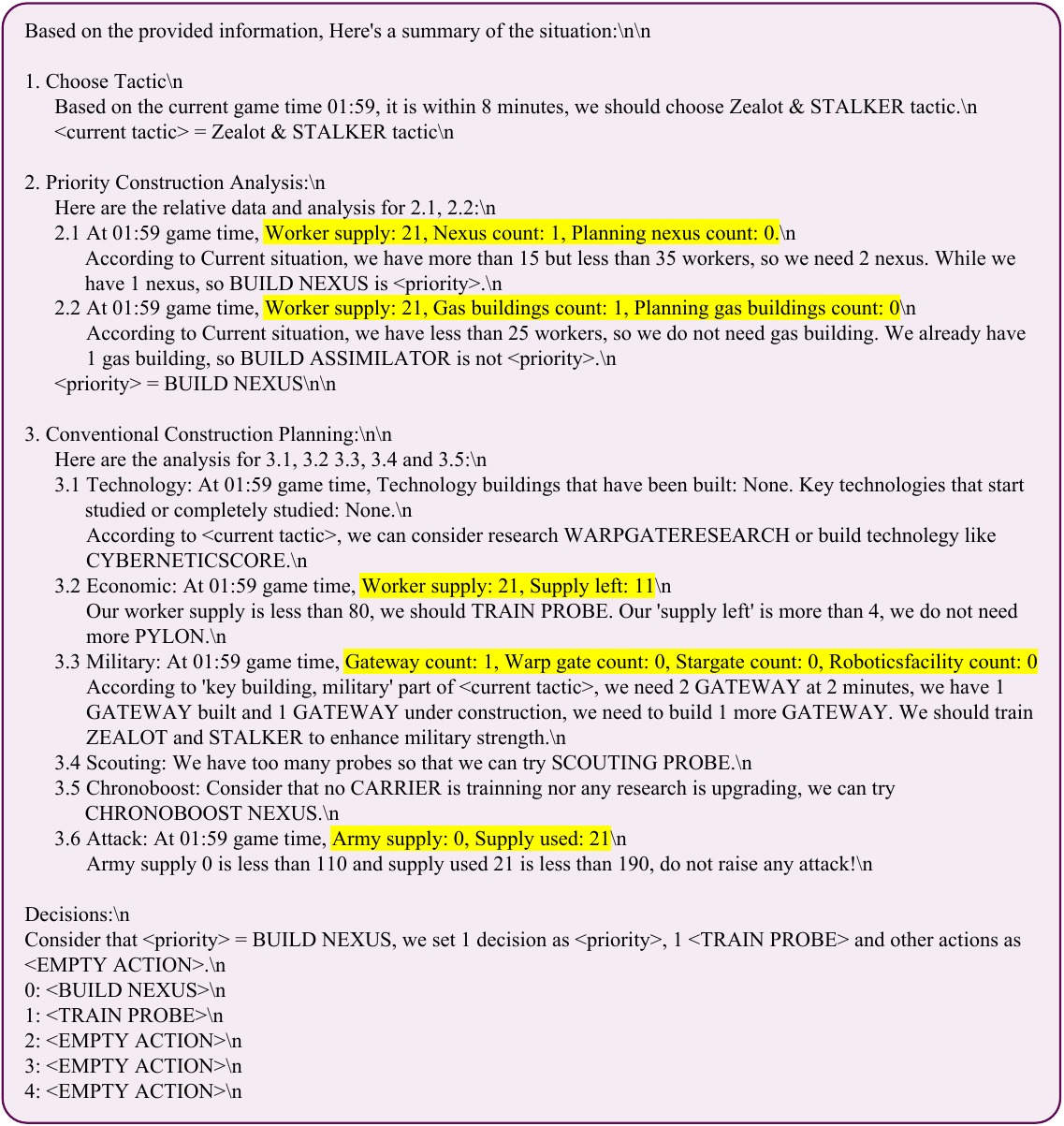}}
\caption{\textbf{Data Extraction in Example output} In the example output, we always list relevant data before giving reasoning results. Importantly, the form of game data should be consistent with the form in observation.} 
\end{figure}

\clearpage
\subsection*{C.3 Never describe conclusion before analyze}

Sometimes in our daily writing, we first write conclusions and then provide analysis. However, describing conclusions before analysis can easily lead to serious reasoning errors when interacting with LLM. So we advise never describe conclusions before analyzing when interact with LLM.

\begin{figure}
\centerline{\includegraphics[width=\textwidth]{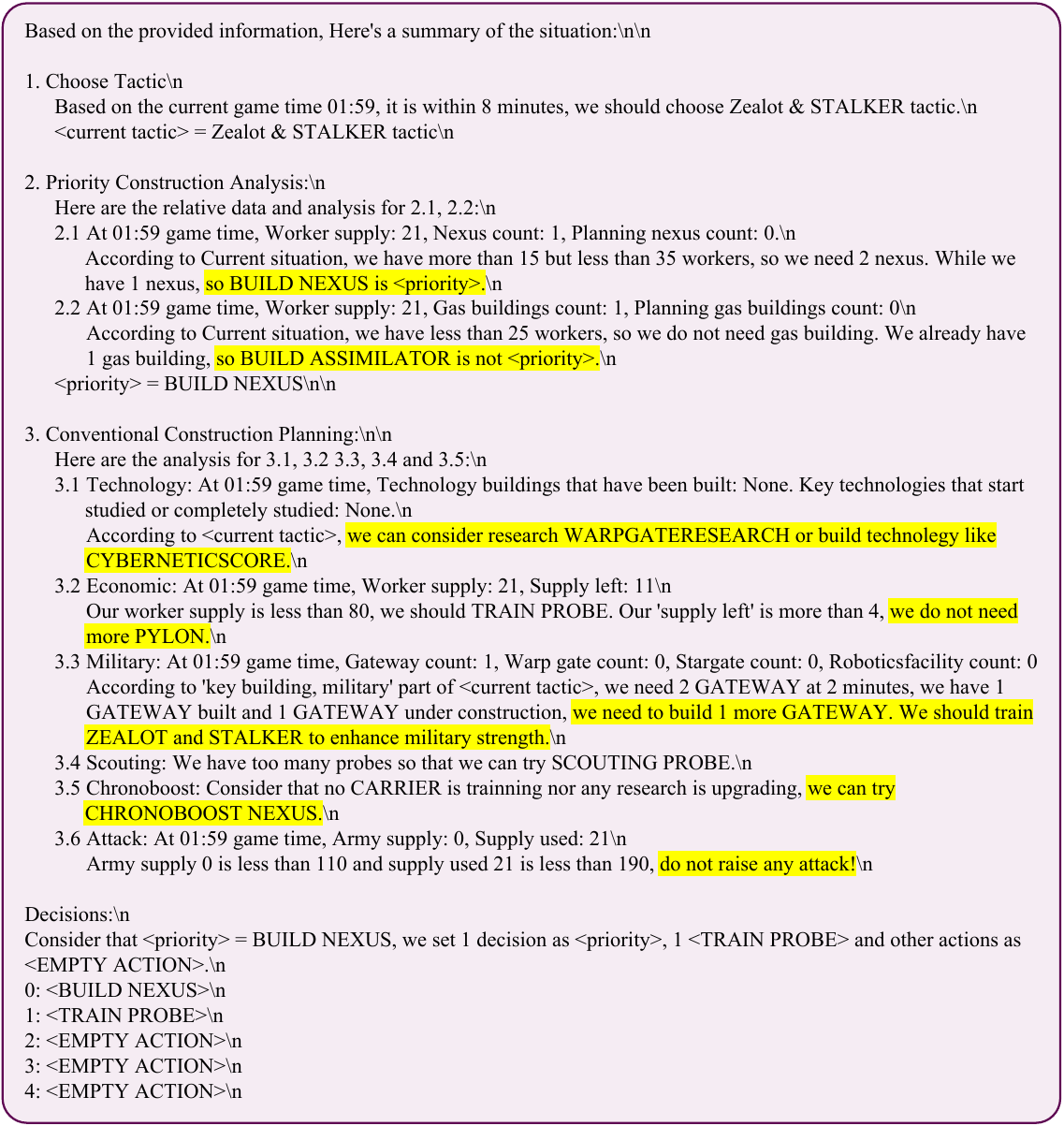}}
\caption{\textbf{Conclusions in Example output} In the example output, we always analyse before giving conclusion/suggestions. First reasoning then give the conclusion is more suitable for LLM inference.} 
\end{figure}


\end{document}